\documentclass{article}

\usepackage{microtype}
\usepackage{graphicx}
\usepackage{subcaption}
\usepackage{booktabs} 

\usepackage{hyperref}

\usepackage[preprint]{icml2026}

\usepackage{amsmath}
\usepackage{amssymb}
\usepackage{mathtools}
\usepackage{amsthm}

\usepackage[capitalize,noabbrev]{cleveref}

\theoremstyle{plain}

\theoremstyle{definition}

\theoremstyle{remark}

\usepackage{multirow}
\usepackage{multicol}
\usepackage{graphicx}
\usepackage{tabularx}
\usepackage{makecell}
\newcolumntype{Y}{>{\centering\arraybackslash}X}

\usepackage{kotex}
\definecolor{mydarkblue}{rgb}{0,0.08,0.45}

\usepackage{pifont}
\newcommand{\cmark}{\ding{51}}  
\newcommand{\xmark}{\ding{55}}  

\icmltitlerunning{RoAD: Robust Autonomous driving under Dataset shifts}

\begin{document}

\twocolumn[
  \icmltitle{RoAD Benchmark: How LiDAR Models Fail \\ under Coupled Domain Shifts and Label Evolution}



  \icmlsetsymbol{equal}{*}

  \begin{icmlauthorlist}
    \icmlauthor{Subeen Lee}{yyy}
    \icmlauthor{Siyeong Lee}{comp}
    \icmlauthor{Namil Kim}{comp}
    \icmlauthor{Jaesik Choi}{sch}
  \end{icmlauthorlist}

  \icmlaffiliation{yyy}{LG AI Research, work done at NAVER LABS while a graduate student at KAIST.}
  \icmlaffiliation{comp}{NAVER LABS}
  \icmlaffiliation{sch}{KAIST}

  \icmlcorrespondingauthor{Jaesik Choi}{jaesik.choi@kaist.ac.kr}

  \icmlkeywords{Computer Vision, 3D Perception, Autonomous Driving, Point Cloud, Transfer Learning, Continual Learning}

  \vskip 0.3in
]



\printAffiliationsAndNotice{}  

\begin{abstract}
For 3D perception systems to operate reliably in real-world environments, they must remain robust to evolving sensor characteristics and changes in object taxonomies.
However, existing adaptive learning paradigms struggle in LiDAR settings where domain shifts and label-space evolution occur simultaneously.
We introduce \textbf{Robust Autonomous Driving under Dataset shifts (RoAD)}, a benchmark for evaluating model robustness in LiDAR-based object classification under intertwined domain shifts and label evolution, including subclass refinement, unseen-class insertion, and label expansion.
RoAD evaluates three learning scenarios with increasing adaptation, from fixed representations (zero-shot transfer and linear probing) to sequential updates (continual learning).
Experiments span large-scale autonomous driving datasets, including Waymo, nuScenes, and Argoverse2.
Our analysis identifies central failure modes:
(i) \textit{limited transferability} under subclass refinement and unseen-class insertion, and on non-vehicle class; and
(ii) \textit{accelerated forgetting during continual adaptation}, driven by feature collapse and self-supervised learning objectives.
\end{abstract}    
\section{Introduction}
\label{sec:intro}

\begin{figure}[t]
  \begin{center}
    \includegraphics[width=\columnwidth]{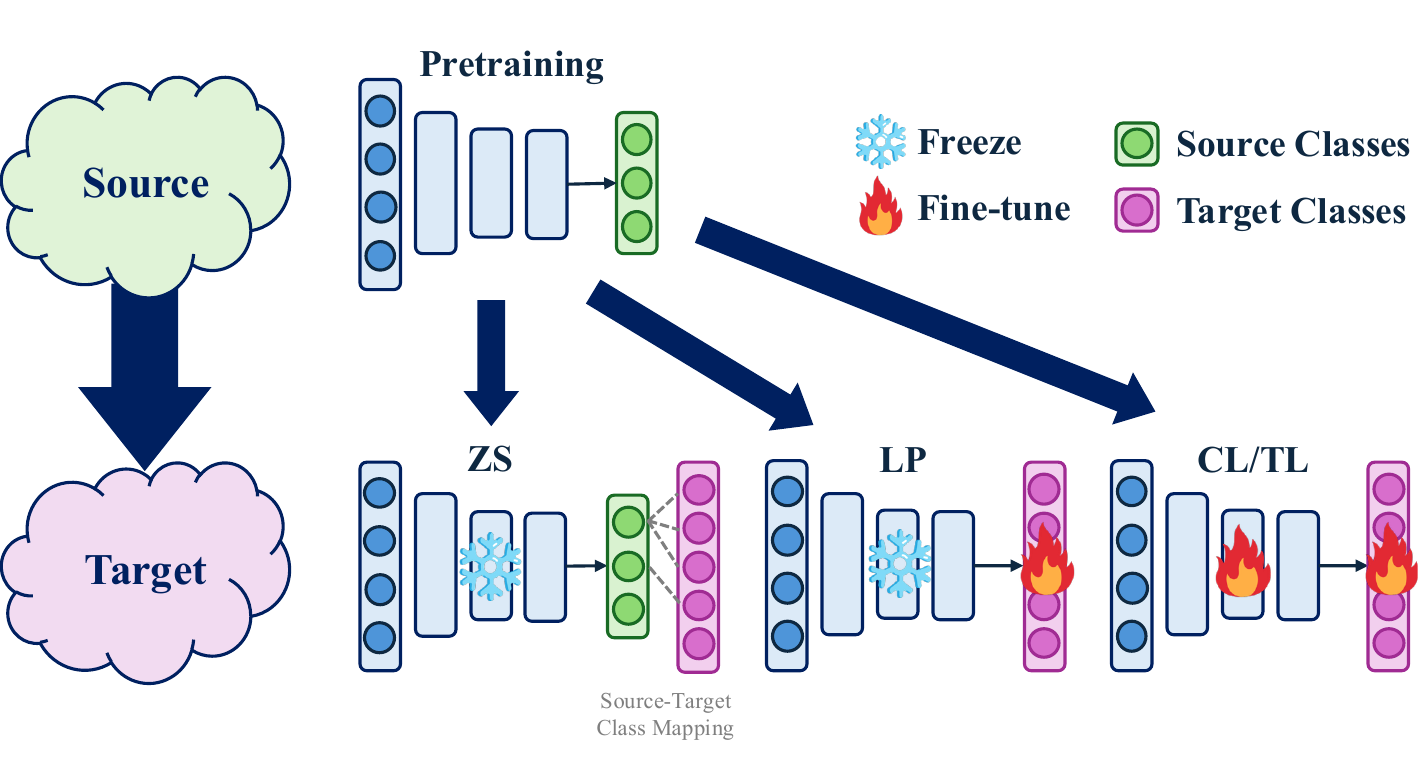}
    \caption{
    Overview of \textbf{RoAD}, a benchmark for robust 3D knowledge transfer under intertwined domain and label shifts, evaluated across zero-shot transfer (ZS), linear probing (LP), and continual learning (CL) from source to target.
    }
    \label{fig:transfer}
  \end{center}
\end{figure}
\begin{figure*}[t]
    \begin{center}
    \begin{subfigure}{0.3\linewidth}
    \includegraphics[width=\linewidth]{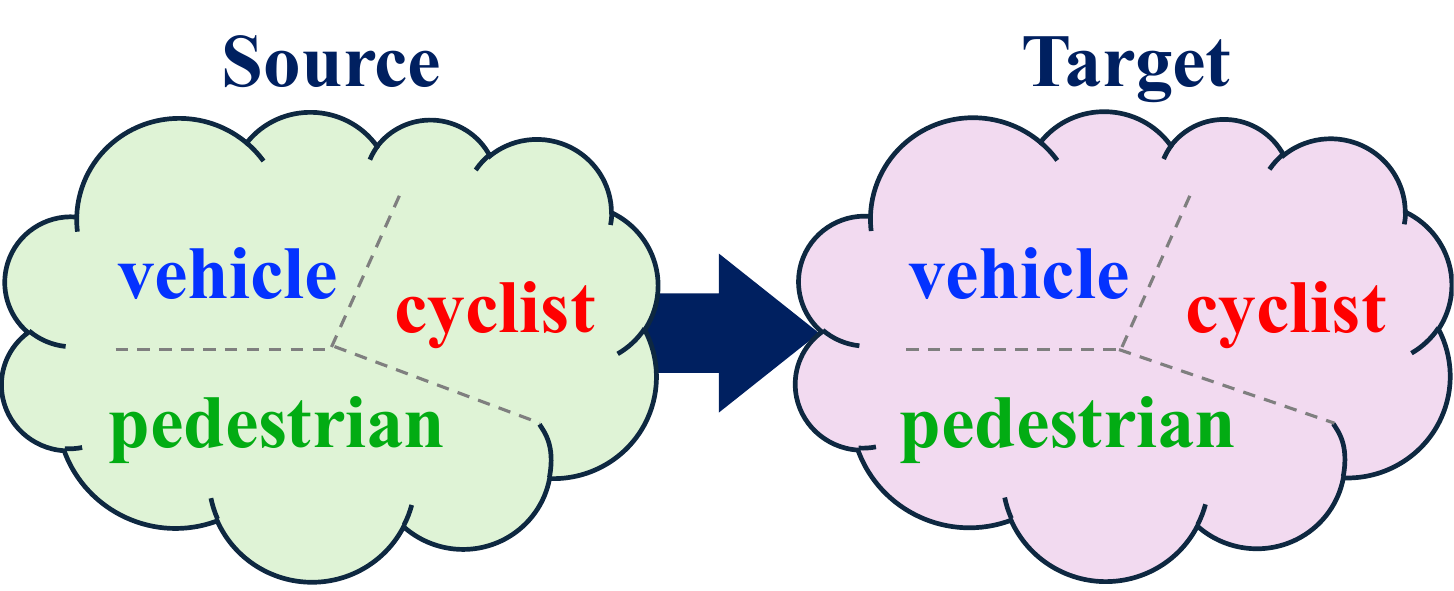} 
        \caption{\textit{Domain shift.}}
    \label{fig:shifts_a}
        \end{subfigure}\hfill
    \begin{subfigure}{0.39\linewidth}
\includegraphics[width=\linewidth]{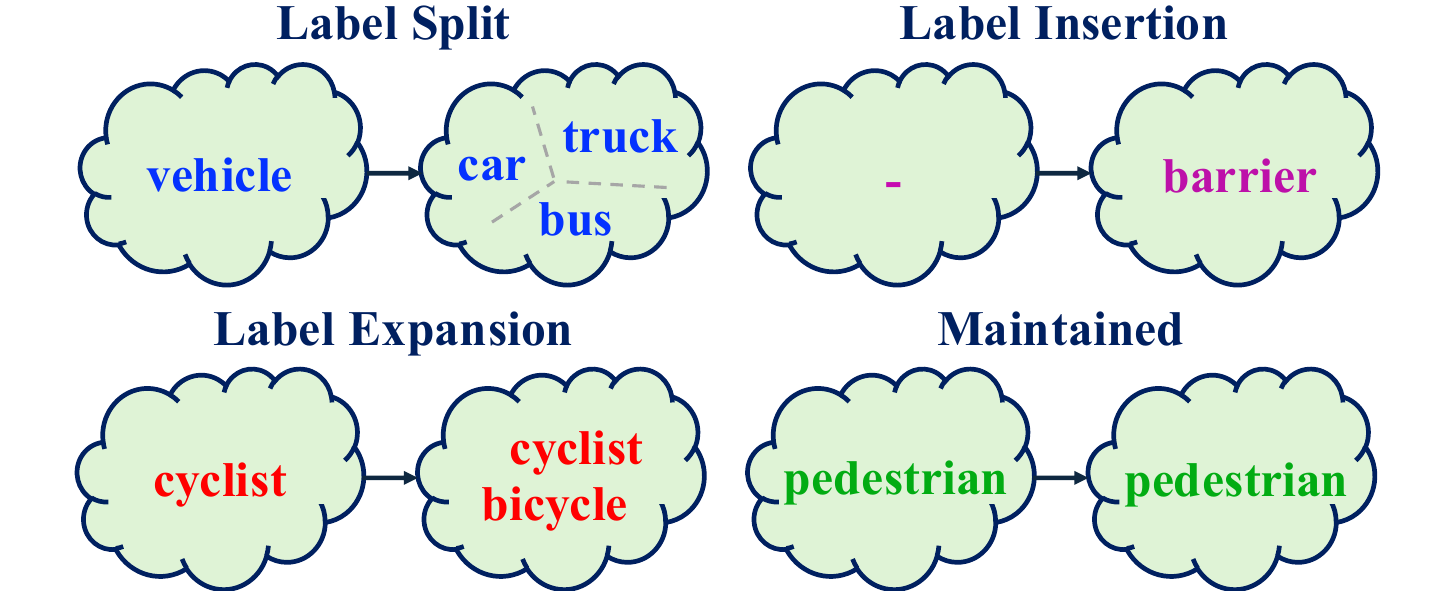}
    \caption{\textit{Types of label shift.}}
\label{fig:shifts_b}
    \end{subfigure}
    \begin{subfigure}{0.3\linewidth}
\includegraphics[width=\linewidth]{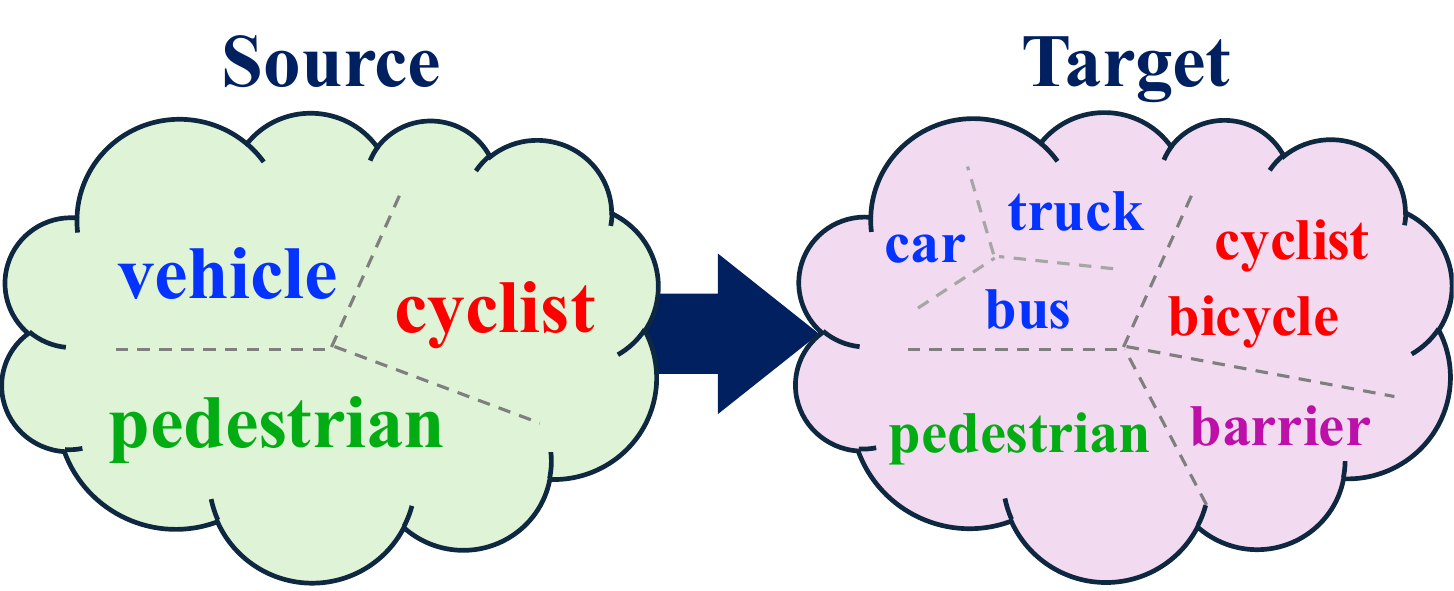}
    \caption{\textit{Coupled shifts (Our setting).}}
\label{fig:shifts_c}
    \end{subfigure}
    \caption{Illustration of domain and label shifts in 3D perception transfer (Waymo$\rightarrow$nuScenes)}
    \label{fig:shifts}
    \end{center}
\end{figure*}

Autonomous driving systems must operate in continually changing environments—such as new cities, varying weather conditions, sensor configurations, and evolving label taxonomies—where collecting and annotating large-scale 3D data is costly.
In such settings, deep learning models must remain robust under distribution and label-space shifts, without relying on expensive retraining from scratch.
This motivates learning paradigms that can efficiently adapt to new knowledge, retain previously learned knowledge, and generalize across diverse data distributions \citep{wang2024comprehensive}.
Across many domains, approaches such as zero-shot learning \citep{wang2019survey}, few-shot learning \citep{wang2020generalizing}, domain adaptation \citep{wang2018deep}, and continual learning \citep{wang2024comprehensive} have been actively explored.

However, knowledge transfer in 3D perception is particularly challenging. First, training and evaluation data are considerably more limited than in 2D vision—a field in which large-scale pretraining has become commonplace.
Second, source-to-target transfer is often hindered by distribution (domain) shifts caused by sensor discrepancies, environmental variation, and geographic diversity \citep{wang2023ssda3d, eskandar2024empirical, zhang2023resimad}.
Moreover, datasets may differ not only in input distributions but also in their label spaces, due to heterogeneous annotation protocols and varying levels of label granularity \citep{Sun_2020_CVPR, caesar2020nuscenesmultimodaldatasetautonomous, DBLP:journals/tmlr/GuoAZDLR25}.
Despite these practical challenges, transfer in 3D perception remains relatively underexplored compared to other vision domains.
Most existing studies assume a fixed label space and focus on 3D object detection, often restricted to a single class (e.g., vehicles) or a few dominant classes \citep{wang2023ssda3d, DBLP:conf/nips/ChangLKKLJJK24, eskandar2024empirical, zhang2023resimad}.
As a result, existing benchmarks are ill-suited for realistic settings where domain gaps and label-space evolution co-occur, creating intertwined shifts that substantially complicate transfer.

In contrast, recent studies in 2D vision have begun to relax such assumptions by proposing more realistic continual learning settings \citep{aljundi2019task, buzzega2020dark, bang2021rainbow, koh2021online, simon2022generalizing, xie2022general, lee2023online}.
These works provide useful templates for evaluating models under evolving label spaces.
In this work, we focus on this largely overlooked setting, which is common in practice yet rarely benchmarked in 3D perception.

Particularly, we study transfer in 3D perception under intertwined domain and label shifts, using a broader set of object categories beyond vehicles.
To better isolate class-level semantic transfer, we focus on object classification rather than detection. 
This design choice is further motivated by two key properties of 3D perception in autonomous driving.
First, unlike 2D image detection where object size varies with camera distance, 3D detection operates in metric scale, making object dimensions strongly correlated with class identity---potentially biasing classification results ~\citep{DBLP:conf/eccv/AliAZZS18}.
Second, objects may be correctly localized yet misclassified, leading to severe downstream decision errors~\citep{DBLP:journals/access/CeccarelliM23}.
In addition, we restrict the input modality to LiDAR point clouds only, removing confounding effects from multimodal fusion.

Building on this formulation, we propose a benchmark named \textbf{RoAD} that systematically characterizes common source-to-target dataset shifts in 3D perception.
Unlike most prior benchmarks that primarily focus on distribution shifts under a fixed label space, RoAD explicitly incorporates label-space evolution as a first-class factor, capturing three types of label space shifts—\textbf{split, insertion, and expansion}—in addition to domain shifts (\cref{fig:shifts}).
The benchmark provides a comparative evaluation framework spanning zero-shot transfer, linear probing, and continual learning.
We instantiate the benchmark using widely adopted perception datasets—Waymo Open Dataset (Waymo) \citep{Sun_2020_CVPR}, nuScenes \citep{caesar2020nuscenesmultimodaldatasetautonomous}, and Argoverse2 \citep{DBLP:conf/nips/WilsonQALSKPKHP21}—and conduct extensive experiments across a range of design choices to provide practical insights for the 3D perception community.
\\

Our main contributions are:
\begin{itemize}
    \item \textbf{RoAD benchmark.} We propose a benchmark for robust knowledge transfer in 3D perception that systematically characterizes \textbf{intertwined domain and label-space shifts}, and evaluates LiDAR-only multi-class object classification across \textbf{zero-shot}, \textbf{linear probing}, and \textbf{continual learning} settings.
    \item \textbf{Instantiation and extensive evaluation.} We instantiate RoAD on Waymo, nuScenes, and Argoverse2, and conduct extensive experiments that reveal practical challenges and key design considerations for transfer under realistic shifts.
    \item \textbf{Empirical findings.} We observe: (1) \textit{zero-shot cyclist transfer fails under coupled shifts}, (2) \textit{label split and insertion hinder linear probes}, (3) \textit{feature collapse accelerates forgetting}, and (4) \textit{supervised learning outperforms self-supervised ones in continual learning}.
\end{itemize}
\section{RoAD: A Benchmark for Robust Autonomous driving under Dataset shifts}
\label{sec:benchmarking}

\subsection{Dataset Shift under Evolving Label Taxonomies}
\subsubsection{Definitions}
In this section, we characterize common dataset shifts encountered when transferring 3D perception models from a source to a target domain.
Dataset shift arises from two sources: \textbf{domain shift} and \textbf{label shift}.
Formally, let $\mathbb{D}$ denote a data distribution and $\mathcal{C}$ a set of class labels.
Consider a source dataset $\mathcal{D}_S$ of samples drawn from $\mathbb{D}_S$ with labels in $\mathcal{C}_S$, and a target dataset $\mathcal{D}_T$ of samples drawn from $\mathbb{D}_T$ with labels in $\mathcal{C}_T$.
We define \textbf{domain shift} (also referred to as \textit{distribution shift} \citep{DBLP:journals/iet-ipr/MadadiSNHM20, DBLP:conf/nips/AnCDH22}) as $\mathbb{D}_S \neq \mathbb{D}_T$, and \textbf{label shift} as $\mathcal{C}_S \neq \mathcal{C}_T$.

We categorize label shifts into three types under label-space expansion: \textbf{label split}, \textbf{label insertion}, and \textbf{label expansion}.
Conversely, these shifts can also be defined in the opposite direction under label-space contraction, corresponding to label merge, label deletion, and label reduction, respectively.
Additionally, some classes may remain unchanged (\textbf{maintained}).
Different shifts capture distinct ways in which semantic knowledge evolves across datasets, as illustrated in \cref{fig:shifts_b}.

\begin{itemize}
    \item \textbf{Label split}: This type relates to hierarchical classification, as it splits coarse classes into finer-grained subclasses. Unlike standard hierarchical classification with a fixed taxonomy, real-world label splits occur under an evolving hierarchy, where future refinements are uncertain and cannot be anticipated \citep{lee2023online,muralidhara2024cleo}. This creates a key challenge: models trained on coarse labels must learn representations rich enough to preserve latent fine-grained distinctions that may later surface as new subclasses.

    \item \textbf{Label insertion}: New classes commonly emerge in real-world applications when new concepts appear, or previously unclassified objects must be recognized. For example, electric scooters—now a prevalent mode of personal transportation—may be added as a new category as data accumulates over time. While this setting is often studied in class-incremental learning (CIL), in practice it often co-occurs with domain shifts \citep{Tan024}, since new data is typically collected in different environments, sensor setups, or geographic regions.

    \item \textbf{Label expansion}: Even when a class label is shared between source and target datasets, its underlying definition can differ. We define label expansion as the case where the target definition strictly subsumes the source definition—i.e., the target class covers a broader set of instances than the corresponding source class.
\end{itemize}

\subsubsection{Examples in 3D Perception Datasets}
This section illustrates concrete examples of domain and label shifts in LiDAR-based 3D perception.
In LiDAR-based 3D perception, commonly discussed sources of \textbf{domain shift} include heterogeneous sensor configurations, differences in LiDAR beam patterns, variations in object scale (e.g., bounding box size distributions), and changing environmental conditions \citep{wei2022lidardistillationbridgingbeaminduced, eskandar2024empirical, zhang2023resimad}.
We summarize the key characteristics of widely used benchmark datasets in \cref{tab:domain_shifts} to highlight how these factors differ across datasets.

In addition to domain shifts, these datasets also differ substantially in their label spaces.
Although many prior works simplify evaluation by transferring from Waymo using only the three shared classes (\texttt{vehicle}, \texttt{pedestrian}, \texttt{cyclist}) \citep{eskandar2024empirical, zhang2023resimad}, label-space evolution naturally arises when transferring from Waymo to nuScenes or Argoverse2.
We summarize representative examples of label shifts in \cref{tab:label_shifts_examples}.
Following OpenPCDet \citep{openpcdet2020} and the official annotations \citep{Sun_2020_CVPR, caesar2020nuscenesmultimodaldatasetautonomous, DBLP:conf/nips/WilsonQALSKPKHP21}, we summarize the full set of class labels in these datasets in \cref{tab:dataset_descript}.


\begin{table}[t]
\caption{Descriptions of Waymo~\citep{Sun_2020_CVPR}, NuScenes ~\citep{caesar2020nuscenesmultimodaldatasetautonomous}, and Argoverse2~\citep{DBLP:conf/nips/WilsonQALSKPKHP21} datasets. $(l,w,h)$ refers to the size of a bounding box.}\label{tab:domain_shifts}
\centering
\resizebox{\linewidth}{!}{
\begin{tabular}{lccc}
\toprule
 & \textbf{Waymo} & \textbf{NuScenes} & \textbf{Argoverse2} \\
\midrule
\midrule
\textbf{LiDAR Sensor} & \begin{tabular}[c]{@{}c@{}}Laser Bear\\Honeycomb\end{tabular} & \begin{tabular}[c]{@{}c@{}}Velodyne\\HDL-32E\end{tabular} & \begin{tabular}[c]{@{}c@{}}Two Velodyne\\VLP-32C\end{tabular} \\
\midrule
\textbf{Site} & \begin{tabular}[c]{@{}c@{}}Phoenix,\\San Francisco,\\Mountain View\end{tabular} & \begin{tabular}[c]{@{}c@{}}Boston,\\Singapore\end{tabular} & \begin{tabular}[c]{@{}c@{}}Austin, Detroit,\\ Miami,\\Pittsburgh,\\Palo Alto,\\Washington, D.C\end{tabular} \\
\midrule
\textbf{Vertical FOV} & $95^\circ$ & $40^\circ$ & $40^\circ$ \\
\midrule
\textbf{\# Beams} & 64 & 32 & 64 \\
\midrule
\textbf{Avg. $(l,w,h)$} & (3.5, 1.6, 1.8)m & (4.9, 2.1, 2.0)m & (4.3, 1.8, 1.9)m \\
\midrule
\textbf{\begin{tabular}[c]{@{}c@{}}Avg. \# Points\\per Object\end{tabular}} & 310 & 75 & 177 \\
\midrule
\midrule
\textbf{\# of labels} & 3 & 10 & 22 \\
\bottomrule
\end{tabular}
}
\end{table}
\begin{table}[t]
\caption{Examples of label shifts when transferring from Waymo to nuScenes/Argoverse2.}\label{tab:label_shifts_examples}
\centering
\resizebox{\linewidth}{!}{
\begin{tabular}{lllr}
\toprule
\textbf{Shift Type} & \textbf{Source} & \textbf{Target} & \textbf{Example} \\
\midrule
\textbf{Split} 
& Waymo & nuScenes
& \texttt{vehicle} $\rightarrow$ \texttt{car}, \texttt{truck}, etc \\
& Waymo & Argoverse2
& \texttt{vehicle} $\rightarrow$ \texttt{car}, \texttt{truck}, etc \\
& Waymo & Argoverse2
& \texttt{pedestrian} $\rightarrow$ \texttt{pedestrian},\\
& & & \hspace*{2em}\texttt{wheeled\_rider} \\
\midrule
\midrule
\textbf{Insert} 
& Waymo & nuScenes
& (-) $\rightarrow$ \texttt{barrier}, \texttt{traffic\_cone} \\
& Waymo & Argoverse2
& (-) $\rightarrow$ \texttt{bollard}, \texttt{sign}, etc \\
\midrule
\textbf{Expand} 
& Waymo & nuScenes
& \texttt{cyclist} $\rightarrow$ \texttt{bicycle} \\
\midrule
\textbf{Maintain} 
& Waymo & nuScenes
& \texttt{pedestrian} $\rightarrow$ \texttt{pedestrian} \\
& Waymo & Argoverse2
& \texttt{cyclist} $\rightarrow$ \texttt{bicyclist} \\
\bottomrule
\end{tabular}
}
\vskip -0.1in
\end{table}

\subsection{Knowledge Transfer Tasks under Dataset Shift}
\subsubsection{Definitions}
To analyze dataset shifts through the lens of model generalizability, we consider three knowledge transfer tasks—\textbf{zero-shot transfer (ZS)}, \textbf{linear probing (LP)}, and \textbf{continual learning (CL)}—spanning increasing degrees of adaptation: from no target training (ZS) to lightweight adaptation (LP) and full model update (CL), illustrated in \cref{fig:transfer}.
This setup allows us to assess the robustness of transfer across different adaptation budgets under realistic deployment settings, where models must repeatedly reuse prior knowledge without retraining from scratch.

All three tasks consist of a pretraining stage on the source dataset and an adaptation (or fine-tuning) stage on the target dataset.
Let $f_S$ be a pretrained model trained on the source dataset $\mathcal{D}_S$ by minimizing the source loss $\mathcal{L}_S$.
Applying $f_S$ directly to the target dataset $\mathcal{D}_T$ without any further training corresponds to \textbf{zero-shot transfer (ZS)}.
\textbf{linear probing (LP)} fine-tunes only the last linear classification layer on the target dataset $\mathcal{D}_T$ while keeping the rest of the network frozen, by minimizing the target loss $\mathcal{L}_T$.
Finally, \textbf{continual learning (CL)} fine-tunes the entire model on $\mathcal{D}_T$ while incorporating an additional regularization term to mitigate forgetting, optimizing $\mathcal{L}_T + \lambda \mathcal{L}_{reg}$, where $\lambda$ controls the strength.
When $\lambda=0$, it is \textbf{transfer learning (TL)}.
For simplicity, we adopt a two-stage setting from a source domain to a target domain ($\mathcal{D}_S$ $\rightarrow$ $\mathcal{D}_T$).

\subsubsection{Evaluation Metrics}
Let $R_{\mathcal{D},f}$ denote the class-averaged accuracy of model $f$ over all evaluated classes in dataset $\mathcal{D}$, and let $R^{c}_{\mathcal{D},f}$ denote its accuracy on class $c$ in $\mathcal{D}$.

For ZS, since the source and target label spaces may differ ($\mathcal{C}_S \neq \mathcal{C}_T$), we apply label mappings for evaluation: split classes are mapped to their unsplit superclass, expanded classes are mapped to their original class, and inserted classes are excluded from evaluation.
For CL, we report the \textbf{backward transfer rate (BWT)},
defined as $(R_{\mathcal{D}_S,f_T} - R_{\mathcal{D}_S,f_S}) / R_{\mathcal{D}_S,f_S}$,
to quantify catastrophic forgetting on the source domain, and the
\textbf{source--target averaged accuracy (ACC)},
defined as $(R_{\mathcal{D}_S,f_T} + R_{\mathcal{D}_T,f_T}) / 2$,
to measure overall performance after adaptation~\citep{lopez2017gradient}.
The tasks and the metrics for each are summarized in \cref{tab:metrics}.
\begin{table}[t]
\centering
\caption{Knowledge transfer tasks and evaluation metrics.}
\label{tab:metrics}
\resizebox{\linewidth}{!}{
\begin{tabular}{lcclc}
\toprule
\textbf{Task} & \textbf{Metrics} & \textbf{Measures} \\
\midrule
\midrule
Zero-Shot Transfer & $R_{\mathcal{D}_T,f}$ & Generalizability of pretrained representation \\
\midrule
Linear Probing & $R_{\mathcal{D}_T,f}$ & Transferability of frozen features \\
\midrule
\multirow{2}{*}{Continual Learning} & ACC & Plasticity (adaptation to new task) \\
 & BWT & Stability (retention of source knowledge) \\
\bottomrule
\end{tabular}
}
\end{table}

\subsubsection{Design Choices}
We evaluate a range of design choices, as summarized in \cref{tab:design}.
Specifically, we vary the source pretraining loss $\mathcal{L}_S$, the continual learning regularization loss $\mathcal{L}_{reg}$, and the backbone architecture, all of which are known to substantially affect transfer performance \citep{li2017learning, kirkpatrick2017overcoming, DBLP:conf/eccv/KimWSS22, DBLP:conf/cvpr/WuSLRVBS22, DBLP:conf/eccv/LiH22}.


\begin{table}[t]
\caption{Design choices for knowledge transfer tasks. Reg. refers to regularization.}
\centering
\label{tab:design}
\resizebox{\linewidth}{!}{
\begin{tabular}{lcll}
\toprule
\textbf{Design} & \textbf{Tasks} & \textbf{Options} & \textbf{Details} \\
\midrule
\midrule
$\mathcal{L}_S$ & ZS, LP, CL & Supervised & CE \\
& & & DR \\
& & & ARPL \\
& & & SupCon \\
& & Self-supervised & InfoNCE \\
& & & VICReg \\
& & Hybrid & CE + InfoNCE \\
& & & CE + VICReg \\
\midrule
$\mathcal{L}_T$ & LP, CL & Supervised & CE \\
\midrule
$\mathcal{L}_{reg}$ & CL & KD-based & LwF \\
& & Reg.-based & EWC \\
\midrule
Backbone & ZS, LP, CL & Point-based & PointNet++ \\
& & & Point Transformer \\
\bottomrule
\end{tabular}
}
\vskip -0.1in
\end{table}

Since $\mathcal{L}_S$ not only determines in-domain source performance but also shapes the representation space that governs downstream generalization, we consider a diverse set of \textbf{supervised and self-supervised pretraining objectives} beyond standard cross-entropy (CE).
These include dot regression or neural collapse-inducing loss (DR) \citep{yang2022inducing}, adversarial reciprocal points learning (ARPL) \citep{chen2021adversarial}, supervised contrastive learning (SupCon) \citep{khosla2020supervised}, and two widely used self-supervised objectives: InfoNCE \citep{simclr} and VICReg \citep{bardes2021vicreg}. 
We also include \textbf{hybrid} losses that linearly combine CE with self-supervised objectives---CE+InfoNCE (NCECE) and CE+VICReg (VRCE).

In CL, $\mathcal{L}_{reg}$ plays a central role in balancing knowledge acquisition on the target dataset with retention of source knowledge, thereby mitigating catastrophic forgetting.
Our setting involves overlapping and semantically related class sets across source and target datasets; therefore, we adopt two classical and well-established approaches that preserve this assumption: \textbf{Learning without Forgetting (LwF)} \citep{li2017learning} and \textbf{Elastic Weight Consolidation (EWC)} \citep{kirkpatrick2017overcoming}.
While existing point cloud CL methods often rely on rehearsal-based strategies \citep{ DBLP:journals/cviu/ZamorskiSKTZ23, ResaniNJ25} or make strong architecture-specific assumptions \citep{DBLP:journals/tnn/DongCSWLLK23, DBLP:conf/iclr/ResaniN25}, we deliberately adopt methods that are \textbf{architecture-agnostic} and \textbf{do not require access to previous-task data} (or stored source samples), aiming for a general-purpose benchmark.
Concretely, LwF maintains separate classification heads for source and target classes: the source head is trained via knowledge distillation to match the logits of $f_S$ on target data, while the target head is optimized using standard supervised classification.
In contrast, EWC uses a single shared head and constrains updates to parameters that are important for the source task by applying a Fisher-information-based quadratic penalty during fine-tuning.

We assume a classifier with a feature-extraction backbone and a lightweight MLP head.
We denote the penultimate-layer representation space as the feature space.
Following \citet{eskandar2024empirical}, which studied the impact of architectural components of 3D object detectors on domain generalizability, we examine the role of backbone architecture in classification.
We adopt two widely used point-based models—PointNet++ \citep{qi2017pointnet++} and Point Transformer \citep{ZhaoJJTK21}—which represent two canonical design paradigms for point clouds: hierarchical local aggregation and attention-based feature interaction, respectively.

\section{Experiments on Waymo, nuScenes, and Argoverse2}\label{sec:experiment}

\begin{table*}[t]
\centering
\caption{\textbf{Accuracy (\%)} for source dataset validation, zero-shot transfer (\textbf{ZS}), linear probing (\textbf{LP}), and continual learning (\textbf{CL}) under different design choices. The \textbf{Waymo} column reports $R_{S,f_S}$; the \textbf{ZS} and \textbf{LP} columns report $R_{T,f_S}$ and $R_{T,f_T}$, respectively; and the \textbf{CL} columns report \textbf{ACC} and \textbf{BWT} in order. Best/worst models within each column and backbone are highlighted in \textbf{bold}/\textit{italic}.}
{\small
\resizebox{\linewidth}{!}{
\begin{tabularx}{\linewidth}{llY|YYYYYY|YYYYYY}
\toprule
& & \multicolumn{1}{c}{} &
\multicolumn{6}{c}{\textbf{W$\rightarrow$N}} &
\multicolumn{6}{c}{\textbf{W$\rightarrow$A}} \\
\cmidrule(lr){4-9}\cmidrule(lr){10-15}

& & \multicolumn{1}{c}{\multirow{2}{*}{\textbf{Waymo}}} &
\multirow{2}{*}{\textbf{ZS}} &
\multirow{2}{*}{\textbf{LP}} &
\multicolumn{4}{c}{\textbf{CL}} &
\multirow{2}{*}{\textbf{ZS}} &
\multirow{2}{*}{\textbf{LP}} &
\multicolumn{4}{c}{\textbf{CL}} \\
\cmidrule(lr){6-9}\cmidrule(lr){12-15}

& & \multicolumn{1}{c}{} &
 & &
\multicolumn{2}{c}{\textbf{LwF}} & \multicolumn{2}{c}{\textbf{EWC}} &
 & &
\multicolumn{2}{c}{\textbf{LwF}} & \multicolumn{2}{c}{\textbf{EWC}} \\
\midrule
\midrule

\multicolumn{1}{l|}{} & \textbf{CE} & 95.6 & 89.8 & 34.8 & 82.5 & -17.6 & 74.8 & -31.1 & 91.6 & 18.6 & 73.6 & -13.1 & 52.7 & -56.3 \\

\multicolumn{1}{l|}{} & \textbf{DR} & 97.8 & 87.7 & 39.6 & \textit{70.3} & \textit{-45.3} & 79.5 & -26.3 & 91.5 & 16.7 & 71.7 & -20.2 & 57.5 & -45.2 \\

\multicolumn{1}{l|}{} & \textbf{ARPL} & 95.1 & 90.6 & 34.1 & 86.9 & -9.7 & 75.8 & -32.3 & 91.7 & 16.2 & \textbf{78.3} & -4.4 & 39.6 & \textit{-84.2} \\

\multicolumn{1}{l|}{} & \textbf{SupCon} & 95.7 & 86.8 & \textit{27.8} & 86.4 & -9.9 & 73.4 & -31.8 & 88.7 & \textit{14.4} & 75.5 & -5.6 & 50.5 & -54.3 \\

\cmidrule(lr){2-15}
\multicolumn{1}{l|}{} & \textbf{NCE} & 93.0 & 85.6 & 50.6 & 82.4 & -13.4 & 73.4 & -34.8 & 85.5 & 34.5 & \textit{69.9} & \textit{-20.5} & \textbf{61.8} & \textbf{-32.9} \\

\multicolumn{1}{l|}{} & \textbf{NCECE} & \textbf{97.9} & \textbf{91.6} & \textbf{57.6} & 84.0 & -19.0 & \textbf{80.7} & -20.7 & \textbf{93.7} & \textbf{38.1} & 73.4 & -19.1 & 57.2 & -42.9 \\

\multicolumn{1}{l|}{} & \textbf{VR} & \textit{87.4} & \textit{78.1} & 47.3 & 84.5 & \textbf{-6.1} & \textit{71.0} & \textit{-35.2} & \textit{76.9} & 24.6 & 76.0 & \textbf{-3.0} & 48.7 & -62.7 \\

\multicolumn{1}{l|}{\multirow{-8}{*}{\rotatebox{90}{\textbf{PointNet++}}}} & \textbf{VRCE} & 96.9 & 89.2 & 44.1 & \textbf{87.2} & -10.3 & 80.1 & \textbf{-18.2} & 90.7 & 26.5 & 78.2 & -5.7 & 40.2 & -79.4 \\

\midrule
\multicolumn{1}{l|}{} & \textbf{CE} & \textbf{96.5} & 89.0 & 32.7 & \textbf{89.2} & -4.8 & \textbf{80.5} & -19.3 & \textbf{91.5} & 12.3 & 74.1 & -9.3 & \textbf{51.1} & \textbf{-56.6} \\

\multicolumn{1}{l|}{} & \textbf{DR} & 91.7 & 84.4 & 35.9 & \textit{70.8} & \textit{-39.2} & 79.5 & \textbf{-18.3} & 88.2 & 15.2 & 71.4 & -13.2 & 44.9 & -63.7 \\

\multicolumn{1}{l|}{} & \textbf{ARPL} & 95.2 & 88.2 & \textit{30.2} & 84.9 & -11.9 & \textit{67.7} & \textit{-31.8} & 90.2 & 12.0 & \textit{66.8} & \textit{-31.3} & 48.2 & -65.4 \\

\multicolumn{1}{l|}{} & \textbf{SupCon} & 95.8 & 88.5 & 34.0 & 87.2 & -6.9 & 77.5 & -22.5 & 89.8 & \textit{11.2} & 74.5 & -8.4 & 38.7 & \textit{-78.0} \\

\cmidrule(lr){2-15}
\multicolumn{1}{l|}{} & \textbf{NCE} & 91.9 & 85.4 & 52.5 & 87.7 & \textbf{-3.3} & 75.9 & -25.9 & 86.4 & \textbf{31.0} & 74.3 & -5.1 & 44.8 & -65.6 \\

\multicolumn{1}{l|}{} & \textbf{NCECE} & 96.2 & 85.0 & \textbf{53.2} & 87.4 & -6.7 & 76.3 & -28.9 & 87.5 & \textbf{31.0} & \textbf{75.4} & -7.9 & 41.4 & -72.5 \\

\multicolumn{1}{l|}{} & \textbf{VR} & \textit{88.0} & \textit{84.0} & 43.8 & 85.3 & -3.8 & 73.2 & -27.4 & \textit{83.8} & 22.7 & 74.6 & \textbf{-2.0} & 44.5 & -61.5 \\

\multicolumn{1}{l|}{\multirow{-8}{*}{\rotatebox{90}{\textbf{Point Transformer}}}} & \textbf{VRCE} & 95.2 & \textbf{90.3} & 45.4 & 86.0 & -8.2 & 73.9 & -26.2 & 91.4 & 22.9 & 70.7 & -12.5 & 45.2 & -68.8 \\
\bottomrule
\end{tabularx}
}}
\vskip -0.1in
\label{tab:all-summary}
\end{table*}

\subsection{Datasets}
As an instantiation of RoAD, we study knowledge transfer from Waymo \citep{Sun_2020_CVPR} to nuScenes \citep{caesar2020nuscenesmultimodaldatasetautonomous} and Argoverse2 \citep{DBLP:conf/nips/WilsonQALSKPKHP21}, where domain and label shifts co-occur under knowledge expansion. 
We denote Waymo$\rightarrow$nuScenes and Waymo$\rightarrow$Argoverse2 as \textbf{W$\rightarrow$N} and \textbf{W$\rightarrow$A}, respectively.

Waymo contains three coarse classes: \texttt{vehicle}, \texttt{pedestrian}, and \texttt{cyclist}. 
nuScenes refines vehicles into six subclasses, retains \texttt{pedestrian}, expands \texttt{cyclist} to \texttt{bicycle}, and adds two new classes (10 total). 
Argoverse2 further splits vehicles (11 subclasses) and pedestrians (3 subclasses), retains \texttt{bicyclist}, and introduces seven additional classes (22 total).

We extract object-level point clouds using ground-truth 3D bounding boxes and use only 3D coordinates ($xyz$) and intensity as input. 
To reduce domain gaps, we apply per-box $xyz$ normalization to $[-1,1]$, and for Waymo, pseudo-line downsampling from 64 to 32 beams \citep{wei2022lidardistillationbridgingbeaminduced}. 
After discarding samples with fewer than 64 points, the resulting train/val splits contain 1.04M/0.25M samples for Waymo, 0.12M/0.02M for nuScenes, and 1.93M/0.39M for Argoverse2.

\subsection{Model Training}
We use standard data augmentations (flipping, rotation, scaling, point sampling, shuffling) with class-balanced resampling; self-supervised objectives use two augmented views per sample. Models are trained with Adam (lr $=10^{-3}$, batch size $=128$), except ARPL+Point Transformer (lr $=10^{-5}$) for stability.
We pretrain on Waymo for 10 epochs, then adapt to targets by fine-tuning for 10/20 epochs on nuScenes and 1/2 epochs on Argoverse2 for LP/CL (chosen by validation convergence).
For NCECE and VRCE, loss coefficients are set to 0.1 (self-supervised) and 1.0 (cross-entropy).
For CL, the regularization weight $\lambda$ is tuned for LwF and EWC, with parameter importance for EWC estimated via the batched empirical Fisher \citep{carta2023avalanche, zhou2023pycil}.
Additional details are provided in \cref{appendix-sec:implementation}.

\section{Results}

\cref{tab:all-summary} summarizes the benchmarking performance.
Most models achieve high validation accuracy on Waymo, indicating that they are sufficiently trained on the source domain. 
In the ZS and LP settings, PointNet++ with NCECE consistently performs best. In contrast, under CL, we do not observe a clear, consistent trend with respect to either the pretraining objective or the backbone.

The following sections highlight key findings from the extensive experiments and relate them to prior work. 
Unless otherwise noted, results use $\mathcal{L}_S$ = CE and PointNet++ as the backbone; the same trends hold across other design choices. Full tables are provided in \cref{appendix-sec:full_table}.

\subsection{Zero-Shot Transfer Fails for Cyclists Under Intertwined Shifts}
\begin{figure}[b]
  \begin{center}
    \includegraphics[width=\columnwidth]{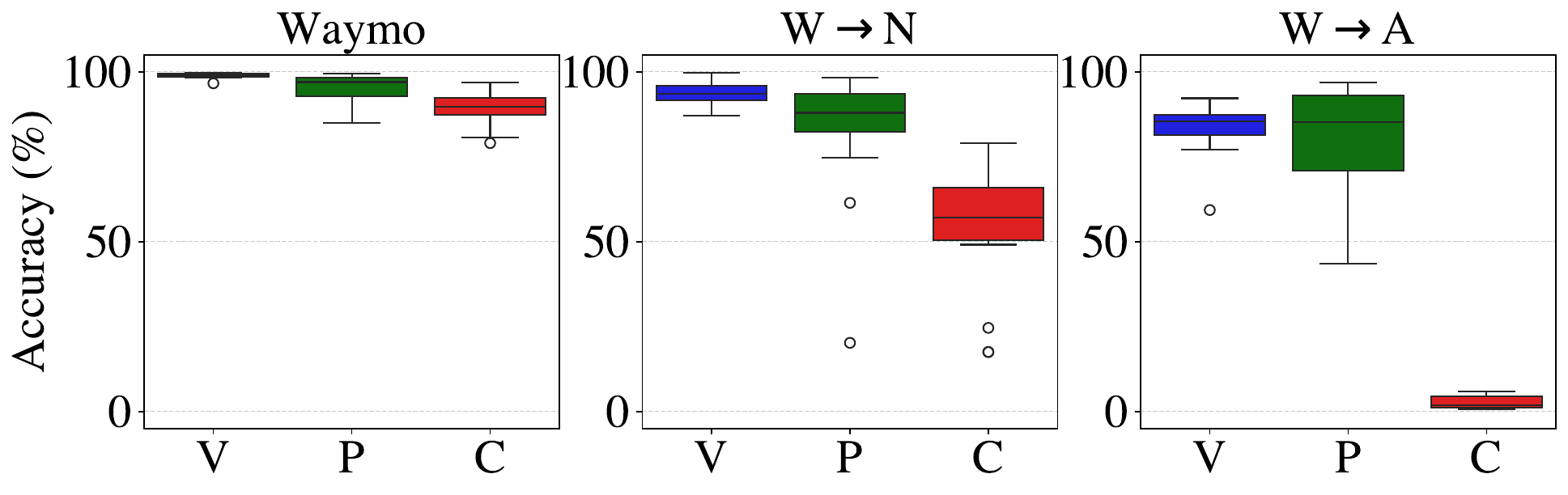}
    \caption{
      \textbf{Distributions of class-wise accuracy} on the Waymo validation set and under \textbf{zero-shot transfer} (W$\rightarrow$N and W$\rightarrow$A), aggregated over all combinations of pretraining objectives and backbone architectures. \textbf{V}, \textbf{P}, and \textbf{C} denote \texttt{vehicle}, \texttt{pedestrian}, and \texttt{cyclist}, respectively.
    }
    \label{fig:zeroshot}
  \end{center}
\end{figure}

In both W$\rightarrow$N and W$\rightarrow$A, \textbf{zero-shot transfer on \texttt{cyclist} consistently underperforms} \texttt{vehicle} and \texttt{pedestrian}, largely independent of the source pretraining loss $\mathcal{L}$ and backbone choice (\cref{fig:zeroshot}).

The underlying cause differs across target datasets. 
In W$\rightarrow$N, the failure is mainly induced by \textbf{label expansion}. 
For nuScenes \texttt{bicycle}, models achieve near-perfect accuracy on \texttt{cyclist} instances (bicycles \emph{with} riders; 100\% for CE+PointNet++), but frequently misclassify bicycles \emph{without} riders (62.3\% for CE+PointNet++). 
This suggests that Waymo-pretrained representations rely on rider-dependent cues, leading to over-specialization that does not capture cyclist-related geometry in a transferable manner.

\begin{figure}[t]
  \begin{center}
    \includegraphics[width=\columnwidth]{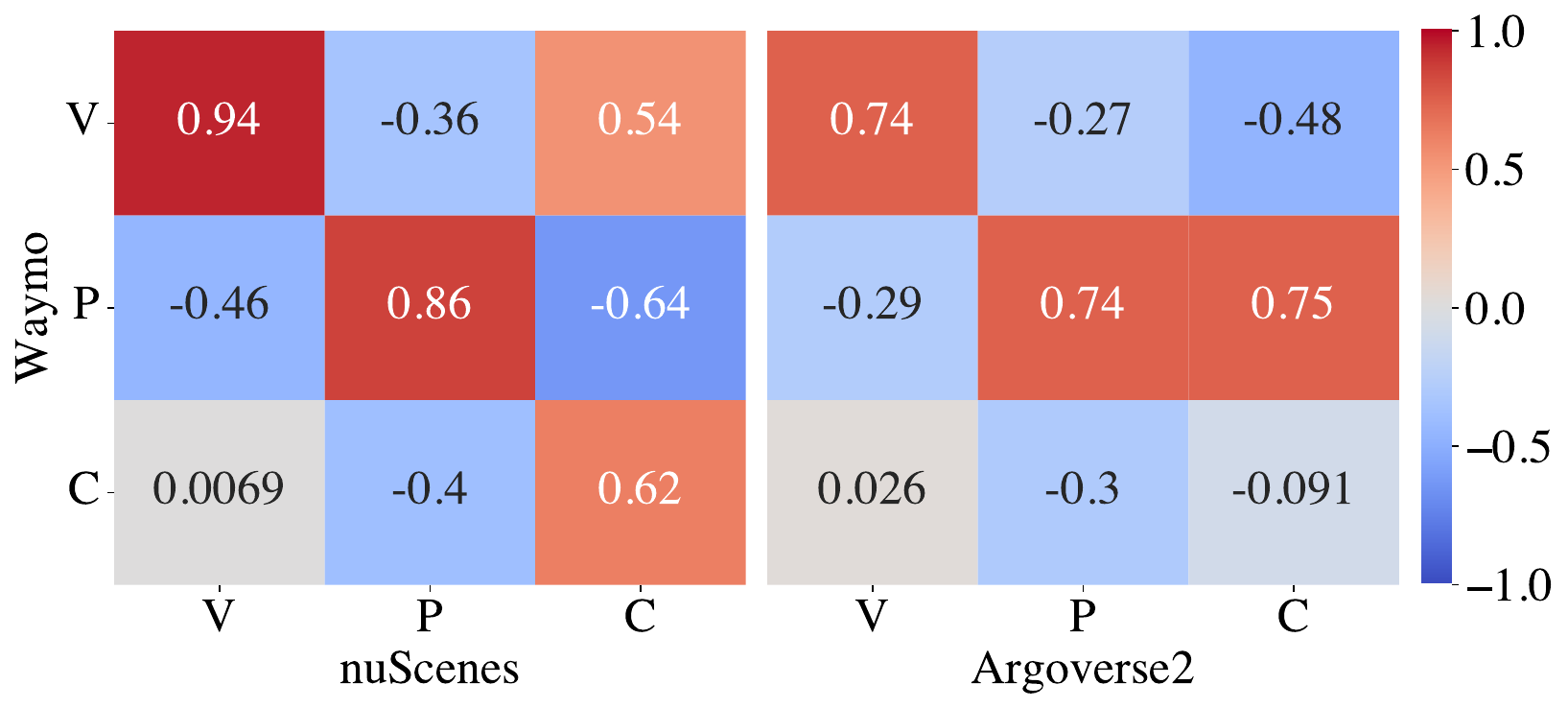}
    \caption{
      \textbf{Cross-dataset class-wise average cosine similarity} on the feature space of the Waymo-pretrained model. \textbf{V}, \textbf{P}, and \textbf{C} denote \texttt{vehicle}, \texttt{pedestrian}, and \texttt{cyclist}, respectively.
    }
    \label{fig:cross-dataset-sim}
  \end{center}
\end{figure}

The failure also persists in W$\rightarrow$A, despite Argoverse2 \texttt{bicyclist} maintaining the Waymo \texttt{cyclist} definition (i.e., label maintained), indicating that \textbf{domain shift} alone can be sufficient to break zero-shot transfer for this class. 
To further analyze this, \cref{fig:cross-dataset-sim} reports cross-dataset class-wise similarity in terms of cosine similarity on the feature space: Argoverse2 \texttt{bicyclist} exhibits very low (even negative) similarity to Waymo \texttt{cyclist}, whereas nuScenes \texttt{bicycle} retains positive similarity to Waymo \texttt{cyclist}. 
This explains why zero-shot transfer remains challenging in W$\rightarrow$A even without label shift.

Overall, these results highlight the need to evaluate transfer beyond vehicles---classes such as cyclists can be substantially more brittle under realistic shifts, yet are often underexplored in prior work. 
Fortunately, the cyclist gap is largely mitigated under more flexible transfer settings (LP and CL).

\begin{figure}[t]
  \begin{center}
    \includegraphics[width=\columnwidth]{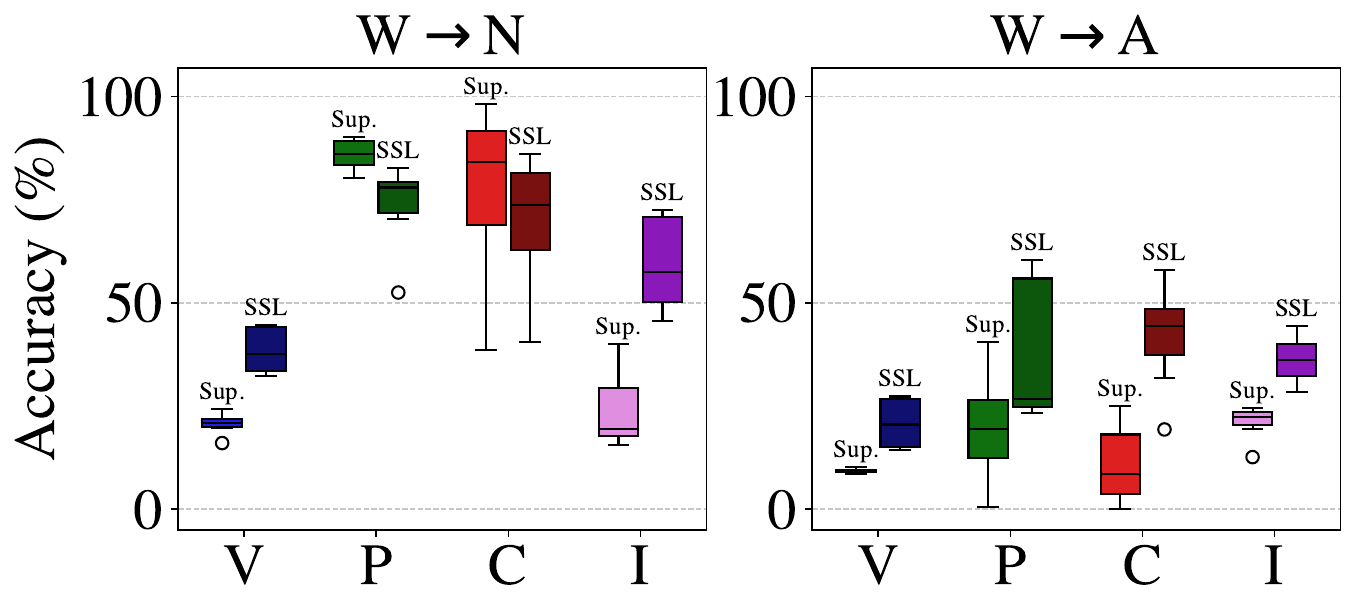}
    \caption{
    \textbf{Distributions of class-wise accuracy} under \textbf{linear probing} (W$\rightarrow$N and W$\rightarrow$A), aggregated across all pretraining objectives and backbone architectures.
    \textbf{Sup.} and \textbf{SSL} denote supervised and self-supervised/hybrid pretraining, respectively.
    \textbf{V}, \textbf{P}, \textbf{C}, and \textbf{I} denote coarse classes \texttt{vehicle}, \texttt{pedestrian}, \texttt{cyclist}, and inserted classes, respectively.
    }
    \label{fig:lp}
  \end{center}
\end{figure}

\subsection{Label Split and Insertion Hinder Transfer}
Under LP, classes affected by \textbf{label split label insertion consistently attain lower accuracy} than classes that are maintained or undergo label expansion, most notably for \texttt{vehicle} and inserted classes in W$\rightarrow$N (\cref{fig:lp}).
Maintained classes or classes undergoing label expansion are typically less disruptive, as they largely preserve the original coarse decision boundaries and require only their refinement, whereas label split and insertion necessitate learning entirely new decision boundaries to separate previously collapsed subclasses or to accommodate novel classes.

\begin{figure}[t]
    \begin{center}
    \includegraphics[width=0.9\columnwidth]{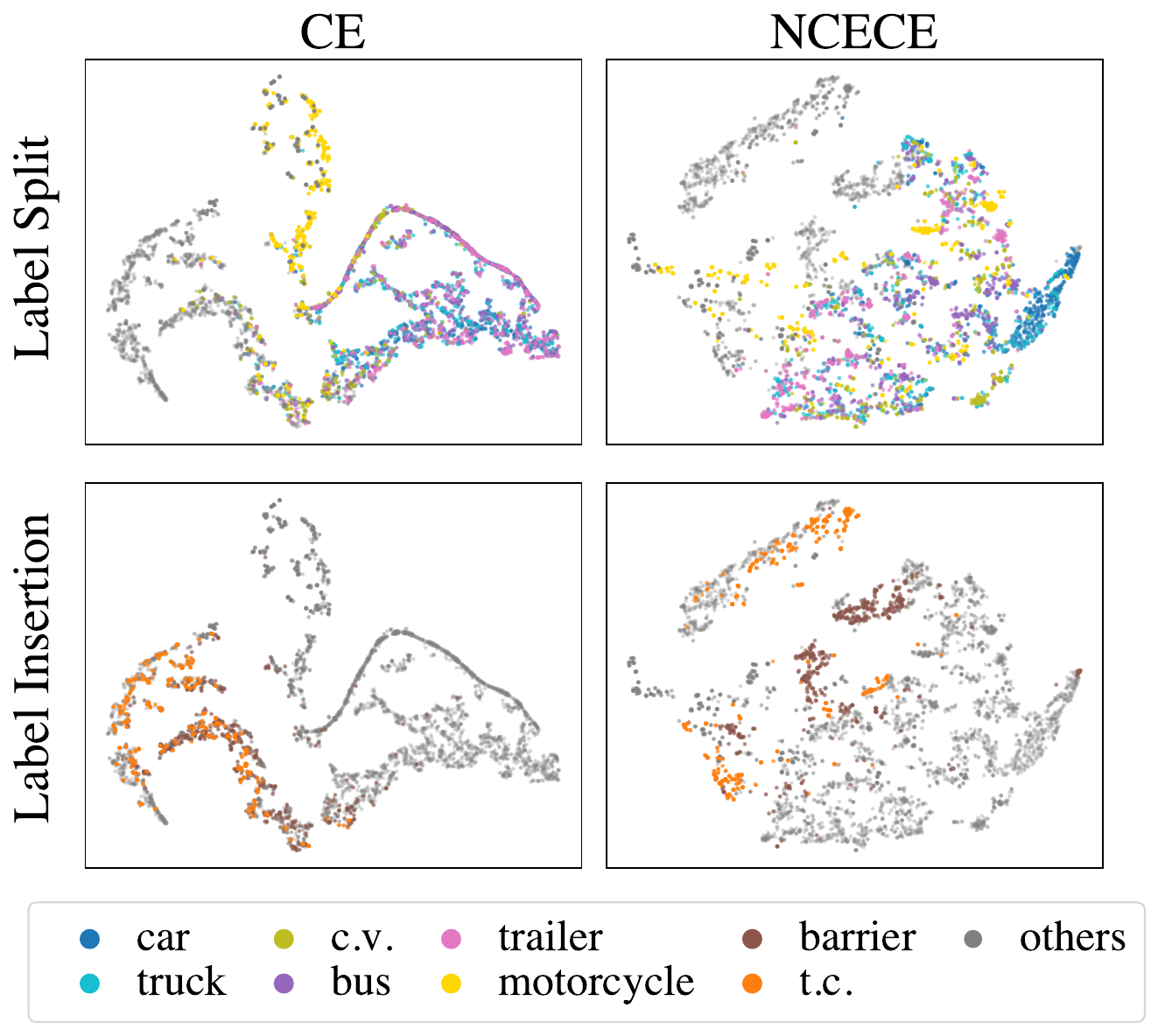}
   \caption{\textbf{t-SNE plots} of label split and insertion classes in nuScenes, using the feature space of $f_S$. c.v. and t.c. refer to \texttt{construction\_vehicle} and \texttt{traffic\_cone}, respectively.}
   \label{fig:tsne_partial}
   \end{center}
\end{figure}

Notably, ZS performs well on coarse superclasses, but accuracy drops sharply when predicting the corresponding fine-grained subclasses. 
This gap suggests insufficient separation of fine-grained classes in the learned representations, consistent with the entangled clusters observed in the t-SNE visualizations (\cref{fig:tsne_partial}). 
Among pretraining objectives, self-supervised and hybrid losses yield improved performance and better feature disentanglement.

These findings motivate reporting transfer performance by label shift type, as different label shifts induce distinct transfer behaviors that may be obscured by aggregate metrics.

\subsection{Feature Collapse Accelerates Forgetting}\label{subsec:dr}
\textbf{Neural-collapse-inducing pretraining (DR) performs worst for CL with LwF} (lowest ACC/BWT) in W$\rightarrow$N.
This is unexpected, as DR has been known to benefit continual and imbalanced learning by mitigating \textit{minority collapse}—where prototypes of minority classes drift toward those of majority classes and become vulnerable to forgetting \citep{yang2022inducing, DBLP:conf/iclr/YangYLLTT23, DBLP:conf/nips/YanQPLZL24}. 
In our setting, however, DR exhibits the most severe catastrophic forgetting.

We attribute this failure to DR’s strong tendency to collapse features toward class prototypes, which preserves only prototype-level information while discarding intra-class variation. 
As a result, when target fine-tuning shifts the feature space, the source information is easily overwritten \citep{dang2025memory}. 
We assume that this phenomenon is amplified when the source and target feature distributions substantially overlap and target fine-tuning induces large updates within this overlapping region.
This mechanism is expected to be most pronounced for CL in \texttt{vehicle} under label split, where the model must separate fine-grained subclasses from features collapsed around a coarse superclass prototype.
To empirically test this hypothesis, we quantify cross-dataset feature overlap on \texttt{vehicle} between Waymo and nuScenes for each pretraining objective and find that DR exhibits the highest overlap among different pretraining objectives, characterized by the highest cosine similarity and second-lowest MMD (\cref{tab:overlap}).

Aligning with recent findings by \citet{DBLP:conf/icml/GraldiBL0N25}, which suggest that stronger feature learning can exacerbate catastrophic forgetting in continual learning, our results indicate a similar trade-off. In our case, a heavily collapsed representation can be effective under the stationary source setting; however, in the non-stationary continual setting, a less-collapsed representation with more conservative (``lazy'') feature updates better mitigates forgetting.

Interestingly, this weakness is largely mitigated under EWC: whereas LwF primarily enforces prototype alignment, EWC constrains parameter updates across layers, enabling retention beyond collapsed prototypes.

\begin{table}[t]
    \centering
    \caption{\textbf{Cross-dataset overlap on \texttt{vehicle} across pretraining objectives} between Waymo and nuScenes, measured by cosine similarity (\textbf{Cos-Sim.}) and maximum mean discrepancy (\textbf{MMD}). Arrows indicate higher overlap. Models with the highest overlap are highlighted in \textbf{bold}.}
    \resizebox{\linewidth}{!}{
    \begin{tabular}{lcclcc}
        \toprule
        \textbf{Objective} & \textbf{Cos-Sim.} $\uparrow$ & \textbf{MMD} $\downarrow$ && \textbf{Cos-Sim.} $\uparrow$ & \textbf{MMD} $\downarrow$ \\
        &  & \small{$\times 10^{-4}$} &&  & \small{$\times 10^{-4}$} \\
        \midrule
        \midrule
        \textbf{CE}     & 0.93 & 3.0    & \textbf{NCE}   & 0.07 & 2.2 \\
        \textbf{DR}     & \textbf{0.99} & 1.2 & \textbf{NCECE} & 0.08 & 6.1 \\
        \textbf{ARPL}   & 0.94 & 254    & \textbf{VR}    & 0.06 & 48 \\
        \textbf{SupCon} & 0.91 & \textbf{0.08} & \textbf{VRCE} & 0.14 & 176 \\
        \bottomrule
    \end{tabular}
    }
    \label{tab:overlap}
    \vskip -0.1in
\end{table}



\subsection{Supervised Pretraining Outperforms Self-Supervised Pretraining in CL}
Although self-supervised and hybrid pretraining objectives outperform supervised pretraining in ZS and LP, this advantage does not persist in CL.
No single pretraining objective or backbone consistently dominates in CL, and in some cases, \textbf{supervised pretraining even outperforms self-supervised or hybrid alternatives.}

This finding appears at odds with prior work \citep{simclr, moco, DBLP:journals/pami/JingT21} suggesting that self-supervised objectives often yield more robust and generalizable representations for adaptation.
We posit that self-supervised learning can provide a suboptimal initialization for CL under domain shift. 
We attribute this to instance-discrimination-based objectives encouraging \emph{domain shortcuts}, where class separation relies on domain-specific cues rather than domain-invariant semantics. 
As CL adapts the feature space to the target domain, these shortcuts are suppressed or removed, which can reduce source-domain performance (i.e., lower backward transfer) because prior decision rules depended on them.

This view aligns with \citet{DBLP:conf/iccv/KimSOPSS21}, who reported that self-supervised models pretrained on non-large-scale source datasets transfer less effectively under domain shifts, whereas large-scale self-supervised pretraining can produce representations that remain discriminative in the target domain.
Likewise, our pretraining is conducted on a single source domain rather than a large, diverse corpus, which may explain its reduced effectiveness in CL.

Following this hypothesis, we expect self-supervised pretraining to remain competitive in CL when domain shift is absent, and only label shift occurs.
We confirm this in nuScenes with coarse-to-fine label refinement and inserted classes---3 coarse classes $\rightarrow$ 6 fine-grained + 2 inserted + 2 maintained classes (10 total)---where self-supervised pretraining achieves comparable ACC but substantially higher BWT (\cref{tab:n3}).

\begin{table}[t]
    \centering
    \caption{\textbf{Accuracy (\%)} for nuScenes 3-class (N3) validation and \textbf{CL from nuScenes 3-class (N3) to nuScenes 10-class (N)} under different pretraining objectives. \textbf{N3} column reports $R_{S,f_S}$; and the \textbf{CL} columns report \textbf{ACC} and \textbf{BWT} in order. Best/worst models within each column are highlighted in \textbf{bold}/\textit{italic}.}
    \begin{tabular}{lccccc}
        \toprule
        \multirow{2}{*}{\textbf{Objective}} &\multirow{2}{*}{\textbf{N3}} & \multicolumn{4}{c}{\textbf{N3$\rightarrow$N (CL)}} \\
        \cmidrule(lr){3-6}
        & & \multicolumn{2}{c}{\textbf{LwF}} & \multicolumn{2}{c}{\textbf{EWC}} \\
        \midrule
        \midrule
        \textbf{CE}  & 97.3 & 90.1 & -1.3 & 86.2 & -0.7 \\
        \textbf{DR}  & 95.8 & \textit{86.6} & \textit{-8.8} & 89.9 & 0.0 \\
        \textbf{ARPL}  & 94.8 & \textbf{93.3} & +7.5 & \textbf{92.6} & +6.7 \\
        \textbf{SupCon}  & \textbf{98.2} & 91.5 & -1.6 & 89.7 & \textit{-3.2} \\
        \midrule
        \textbf{NCE}  & \textit{79.8} & 91.5 & +19.2 & 90.9 & \textbf{+18.9} \\ 
        \textbf{NCECE}  & 93.0 & 91.3 & +14.5 & 80.9 & +9.2 \\
        \textbf{VR}  & 87.4 & 91.0 & +7.7 & 90.0 & +11.8 \\
        \textbf{VRCE}  & 90.1 & 90.0 & \textbf{+21.2} & \textit{74.1} & +9.3 \\
        
        \bottomrule
    \end{tabular}
    \label{tab:n3}
    \vskip -0.1in
\end{table}

\section{Related Work}
\label{sec:related}
\subsection{Continual Learning}
Continual learning (CL) enables models to learn sequential tasks without forgetting prior knowledge, with techniques to prevent catastrophic forgetting \citep{mccloskey1989catastrophic}. 
Learning without Forgetting (LwF) \citep{li2017learning} applies knowledge distillation at the output (logit) level to preserve the structure of the learned feature space.
Regularization-based methods constrain parameter updates to preserve past knowledge, e.g., EWC \citep{kirkpatrick2017overcoming} uses the Fisher Information matrix to estimate the importance of each neuron to regularize.

Recent CL studies in 3D point cloud recognition \citep{abs-2304-03980, eskandar2024empirical, Tan024, ResaniNJ25} largely adopt experimental protocols from the 2D image domain and rarely consider realistic settings, such as label hierarchies or intertwined domain and label shifts.
Furthermore, CL remains underexplored in LiDAR-based perception systems \citep{abs-2304-03980, DBLP:journals/ral/CuiC23}.
Although \cite{Tan024} considers both domain shift and CIL, it assumes dense points ($\geq$1024 points) and disjoint classes across tasks, limiting real-world applicability (such as LiDAR). Our work investigates CL under joint domain and diverse label shifts in LiDAR datasets.

\subsection{3D Point Cloud Recognition}
Deep learning methods for 3D point cloud recognition can be broadly categorized into projection-, voxel-, and point-based approaches.
Projection-based methods project 3D points onto 2D views and apply CNNs \citep{SuMKL15, ChenMWLX17}, often sacrificing geometric fidelity.
Voxel-based methods discretize space for 3D convolutions \citep{MaturanaS15}, with sparse convolutional variants improving efficiency \citep{GrahamEM18, ChoyGS19}, though scalability and resolution trade-offs remain.
Point-based methods directly operate on raw point sets and preserve fine-grained geometric information, making them particularly suitable for LiDAR data \citep{QiSMG17, qi2017pointnet++, ThomasQDMGG19, ZhaoJFJ19}.
Early architectures rely on hierarchical set abstraction and local aggregation \citep{qi2017pointnet++}, while more recent approaches incorporate attention and transformer mechanisms to model long-range dependencies \citep{GuoCLMMH21, ZhaoJJTK21, 0002LJLZ22, 00020WLL0O0Z24}.
Following this line of work, we adopt PointNet++ \citep{qi2017pointnet++} and Point Transformer \citep{ZhaoJJTK21} as representative convolution- and attention-based point architectures in our experiments.

\subsection{Generalizability in Representation Learning}
Generalizable representations are crucial for transfer across domains and tasks. 
Although supervised learning achieves strong task-specific performance, it often degrades under domain shifts or novel label spaces, especially in limited-data regimes where overfitting is more likely.
Self-supervised learning (SSL) leverages unlabeled data and is commonly believed to produce more transferable representations \citep{simclr, moco, DBLP:journals/pami/JingT21}. 
However, representation generalizability is typically evaluated in static settings, such as zero-shot transfer or linear probing \citep{RadfordKHRGASAM21}. 
In this work, we evaluate generalizability under both static and non-static settings to systematically assess the robustness of common design choices.

\section{Conclusion \& Future Directions}
\label{sec:conclusion}
We introduced RoAD, a benchmark for robust knowledge transfer in 3D perception that jointly captures domain and label shifts. 
By evaluating multiple transfer settings across two realistic scenarios---Waymo-to-nuScenes/Argoverse2---RoAD provides actionable insights for building robust LiDAR-based perception systems. 
Beyond autonomous driving, we expect RoAD to inform broader 3D learning research, particularly in settings where point clouds are sparse and noisy.
Future work will also explore richer shift taxonomies (e.g., long-tail and open-set label shifts) and more challenging tasks such as 3D object detection and motion forecasting.

\section*{Impact Statement}
This work studies how LiDAR-based 3D perception models transfer to new dataset when both the domain and the label set change, as in real autonomous driving where sensors, environments, and object definitions evolve over time. 
We evaluate zero-shot transfer, linear probing, and continual learning across Waymo, nuScenes, and Argoverse2 to see when and why performance drops.
From a safety standpoint, we find clear risks: harder classes (e.g., cyclists) often transfer poorly, and some types of label changes lead to more errors and faster forgetting during continual learning. 
These issues can be hard to notice because a model may look strong on familiar validation data but fail after deployment shifts.
By offering a realistic benchmark and analyzing these failure patterns, our work encourages safer evaluation and motivates methods that focus on robustness and stability during ongoing adaptation. 
We hope these insights help researchers and practitioners anticipate transfer failures and reduce safety risks in real-world.

{
    \bibliographystyle{icml2026}
    \bibliography{main}
}

\clearpage
\appendix
\section{Dataset \& Implementation Details}\label{appendix-sec:implementation}
\subsection{Dataset Description}
\cref{tab:dataset_descript} summarizes the class-wise sample counts in the training and validation splits after applying the preprocessing described in the experimental section. 
For source and target class selection, we follow the default OpenPCDet protocol. 
For Argoverse2, we additionally remove four rare classes (\texttt{construction\_cone}, \texttt{dog}, \texttt{message\_board\_trailer}, and \texttt{wheelchair}), each occurring in fewer than 0.02\% of the training set after filtering out objects with fewer than 64 points.

\subsection{Model Training}
All experiments are conducted on an NVIDIA A100 GPU using PyTorch \citep{DBLP:conf/nips/PaszkeGMLBCKLGA19} v1.12.1 with CUDA 11.3. 
We optimize all models using Adam (weight decay $=0.01$, $\beta_1=0.9$; other settings use default values). 
Training schedules are chosen to ensure loss convergence: with a batch size of 128, source pretraining converges in approximately 80K iterations. 
LP fine-tuning requires approximately 9K/15K iterations on nuScenes/Argoverse2, while CL requires approximately 19K/30K iterations.

For ARPL~\citep{chen2021adversarial}, we use a single prototype per class, a margin loss weight of 10.0, and a fixed margin radius of 1.0. 
For self-supervised learning objectives, the penultimate feature layer—i.e., the output after the third fully connected layer—is used to compute the loss. 
For VICReg~\citep{bardes2021vicreg}, loss weights for similarity, standard deviation, and covariance are set to (25.0, 25.0, 1.0), following the original implementation.

For N3$\rightarrow$N discussed in Section~4.4, we reconstruct nuScenes into a 3-class source dataset consisting of \texttt{vehicle}, \texttt{pedestrian}, and \texttt{bicycle}, where \texttt{vehicle} aggregates \texttt{car}, \texttt{truck}, \texttt{construction\_vehicle}, \texttt{bus}, \texttt{trailer}, and \texttt{motorcycle}. 
We refer to this dataset as nuScenes-3 (N3) and denote transfer to the original 10-class nuScenes as N3$\rightarrow$N. 
In this setting, models are pretrained for 20 epochs, fine-tuned for 10 epochs for SupCon/NCE/VR last-layer training and LP, and trained for 20 epochs for CL. 
We use a learning rate of $10^{-3}$ for all models, except for ARPL pretraining, which uses $10^{-4}$ to avoid representation collapse observed at $10^{-3}$. 
All other settings follow W$\rightarrow$N.

\subsection{Model Architectures}
The model architecture consists of a PointNet++ \citep{qi2017pointnet++} or Point Transformer \citep{ZhaoJJTK21} backbone followed by a 4-layer fully connected classifier head with output dimensions of [512, 256, 128, $num\_classes$]. 
Each layer includes Batch Normalization and ReLU activation.

PointNet++ employs a stacked multi-scale grouping (MSG) configuration defined as follows (notation follows the original PointNet++ paper): $SA(K, r, [l_1, ..., l_d])$ denotes a Set Abstraction (SA) level with $K$ local regions of ball radius $r$, using PointNet with $d$ fully connected layers of widths $l_i$. $SA([l_1, ..., l_d])$ represents a global SA level aggregating the entire set into a single vector. 
For multi-scale settings, $SA(K, [r^{(1)}, ..., r^{(m)}], [[l^{(1)}_1, ..., l^{(1)}_d], ..., [l^{(m)}_1, ..., l^{(m)}_d]])$ indicates MSG with $m$ radii and corresponding MLPs.
Our specific architecture for PointNet++ backbone is: 

\begin{table}[h]
\centering
\caption{PointNet++ model architecture.}
\resizebox{\linewidth}{!}{%
\begin{tabular}{lc}
\toprule
Layer & Configuration \\
\midrule
SA 1 & $256,\;[0.1,0.2,0.4],\;[[32,32,64],[64,64,128],[64,96,128]]$ \\
SA 2 & $64,\;[0.2,0.4,0.8],\;[[64,64,128],[128,128,256],[128,128,256]]$ \\
SA 3 & $16,\;1.0,\;[256,512,1024]$ \\
\bottomrule
\end{tabular}%
}
\vskip -0.1in
\label{tab:pn_arch}
\end{table}

Point Transformer employs residual transformer blocks composed of self-attention and linear projection layers, followed by downsampling. 
We stack five encoder blocks with downsampling rates [1,2,2,4,4], producing point set sizes [N, N/2, N/4, N/16, N/64], where $N$ denotes the number of input points. 
All other hyperparameters follow the original implementation \citep{ZhaoJJTK21}.

\begin{table}[b]
\centering
\caption{\textbf{Accuracy (\%)} for source dataset validation and W$\rightarrow$N zero-shot transfer (\textbf{ZS}), linear probing (\textbf{LP}), and continual learning (\textbf{CL}) under different loss coeffcients for hybrid pretraining objectives. \textbf{Coef.} refers to the self-supervised loss coefficient. The \textbf{Waymo} column reports $R_{S,f_S}$; the \textbf{ZS} and \textbf{LP} columns report $R_{T,f_S}$ and $R_{T,f_T}$, respectively; and the \textbf{CL} columns report \textbf{ACC} and \textbf{BWT} in order. Best models within each column and each pretraining objective are highlighted in \textbf{bold}.}\label{tab:ssl_abl}
{\small
\resizebox{\linewidth}{!}{
\begin{tabular}{ll|c|cccccc}
\toprule
& \multicolumn{1}{c}{\multirow{2}{*}{\textbf{Coef.}}} & \multicolumn{1}{c}{\multirow{2}{*}{\textbf{Waymo}}} &
\multirow{2}{*}{\textbf{ZS}} &
\multirow{2}{*}{\textbf{LP}} &
\multicolumn{4}{c}{\textbf{CL}}  \\
\cmidrule(lr){6-9}

& \multicolumn{1}{c}{} & \multicolumn{1}{c}{} &
 & &
\multicolumn{2}{c}{\textbf{LwF}} & \multicolumn{2}{c}{\textbf{EWC}} \\
\midrule
\midrule

\multirow{3}{*}{\textbf{NCECE}}
& \textbf{0.01} & 97.7 & 74.7 & 50.6 &80.0 &-24.2 &75.4 &-33.9 \\
& \textbf{0.1} & \textbf{97.9} & \textbf{91.6} & 57.6 &\textbf{84.0} &\textbf{-19.0} &\textbf{80.7} &\textbf{-20.7} \\
& \textbf{1.0} & 97.7 & 91.4 & \textbf{57.8} &80.3 &-22.4 &79.9 &-23.1 \\
\midrule
\multirow{3}{*}{\textbf{VRCE}} 
& \textbf{0.01} & \textbf{98.1} & 86.9 & 36.2 &77.7 &-28.2 &80.0 &-25.4 \\
& \textbf{0.1} & 96.9 & \textbf{89.2} & \textbf{44.1} &\textbf{87.2} &\textbf{-10.3} &\textbf{80.1} &\textbf{-18.2} \\
& \textbf{1.0} & 92.9 & 88.7 & 42.9 &82.0 &-16.7 &72.8 &-35.6 \\
\bottomrule
\end{tabular}
}
}
\end{table}

\begin{table*}[t]
\caption{\textbf{Class-wise number of Train/Val samples} in Waymo, nuScenes, and Argoverse2 datasets.}
\resizebox{\linewidth}{!}{
\begin{tabular}{ll|cccc}
\toprule
\multicolumn{2}{l|}{\textbf{Dataset}}                                                       & \multicolumn{4}{c}{\textbf{Class Name (Number of Samples)}}                                                                                                                                                                                                                                                \\
\midrule
\midrule
\multicolumn{1}{l|}{\multirow{2}{*}{\textbf{Waymo}}}      & \multicolumn{1}{l|}{\textbf{Train}} & \multicolumn{1}{c|}{\texttt{vehicle}(647,571)}                                                                                                                                       & \multicolumn{1}{c|}{\texttt{pedestrian}(367,319)} & \multicolumn{1}{c|}{\texttt{cyclist}(23,542)} & -                                        \\
\cline{2-6}
\multicolumn{1}{l|}{}                                     & \multicolumn{1}{l|}{\textbf{Val}}   & \multicolumn{1}{c|}{\texttt{vehicle}(161,447)}                                                                                                                                       & \multicolumn{1}{c|}{\texttt{pedestrian}(87,286)}  & \multicolumn{1}{c|}{\texttt{cyclist}(5,910)}  &  -                                       \\ \midrule
\multicolumn{1}{l|}{}   & \multicolumn{1}{l|}{\textbf{Train}} 
& \multicolumn{1}{c|}{\begin{tabular}[c]{@{}c@{}}\texttt{car}(65,025), \texttt{truck}(16,543),\\ \texttt{bus}(4,016), \texttt{trailer}(5,966),\\ \texttt{construction\_vehicle}(2,646),\\ \texttt{motorcycle}(1,128)\end{tabular}} & \multicolumn{1}{c|}{\texttt{pedestrian}(5,117)}   & \multicolumn{1}{c|}{\texttt{bicycle}(433)}    & \texttt{barrier}(18,917), \texttt{traffic\_cone}(1,579) \\
\cline{2-6}
\multicolumn{1}{l|}{\multirow{-6}{*}{\rotatebox{0}{\textbf{nuScenes} }}}                                   & \multicolumn{1}{l|}{\textbf{Val}}   & \multicolumn{1}{c|}{\begin{tabular}[c]{@{}c@{}}\texttt{car}(12,205), \texttt{truck}(3,351),\\ \texttt{bus}(1,031), \texttt{trailer}(1,015),\\ \texttt{construction\_vehicle}(378),\\ \texttt{motorcycle}(220)\end{tabular}}      & \multicolumn{1}{c|}{\texttt{pedestrian}(1,139)}   & \multicolumn{1}{c|}{\texttt{bicycle}(57)}     & \texttt{barrier}(3,537), \texttt{traffic\_cone}(219)    \\ 
\midrule
\multicolumn{1}{l|}{} & \multicolumn{1}{l|}{\textbf{Train}} 
& \multicolumn{1}{c|}{\begin{tabular}[c]{@{}c@{}}\texttt{regular\_vehicle}(1,354,539), \texttt{bus}(47,353),\\ \texttt{large\_vehicle}(45,511), \texttt{box\_truck}(42,816),\\ \texttt{truck}(31,458), \texttt{vehicular\_trailer}(19,485), \\ \texttt{motorcycle}(8,937), \texttt{truck\_cab}(8,717),\\  \texttt{school\_bus}(6,276), \texttt{articulated\_bus}(5,710),\\ \texttt{motorcyclist}(388)\end{tabular}}   
& \multicolumn{1}{c|}{\begin{tabular}[c]{@{}c@{}}\texttt{pedestrian}(280,053),\\ \texttt{wheeled\_rider}(682),\\ \texttt{stroller}(558)\end{tabular}}                     & \multicolumn{1}{c|}{\texttt{cyclist}(3,967)}                 & 
\multicolumn{1}{c}{\begin{tabular}[c]{@{}c@{}}\texttt{stop\_sign}(32,852), \texttt{sign}(20,481), \\ \texttt{bicycle}(15,301),  \texttt{construction\_barrel}(9,038), \\ \texttt{wheeled\_device}(5,249), \texttt{bollard}(3,592),\\ \texttt{mobile\_pedestrian\_crossing\_sign}(329)\end{tabular}}  
\\
\cline{2-6}
\multicolumn{1}{l|}{\multirow{-6}{*}{\rotatebox{0}{\textbf{Argoverse2} }}}                                   & \multicolumn{1}{l|}{\textbf{Val}} & \multicolumn{1}{c|}{\begin{tabular}[c]{@{}c@{}}\texttt{regular\_vehicle}(282,696), \texttt{bus}(7,229),\\ \texttt{large\_vehicle}(7,759), \texttt{box\_truck}(9,716),\\ \texttt{truck}(5,970), \texttt{vehicular\_trailer}(5,974), \\ \texttt{motorcycle}(1,724), \texttt{truck\_cab}(4,241),\\ \texttt{school\_bus}(1,391), \texttt{articulated\_bus}(996)\\ \texttt{motorcyclist}(83)\end{tabular}}   
& \multicolumn{1}{c|}{\begin{tabular}[c]{@{}c@{}}\texttt{pedestrian}(47,634),\\ \texttt{wheeled\_rider}(60),\\ \texttt{stroller}(72)\end{tabular}}                     & \multicolumn{1}{c|}{\texttt{cyclist}(323)}                 & 
\multicolumn{1}{c}{\begin{tabular}[c]{@{}c@{}}\texttt{stop\_sign}(6,263), \texttt{sign}(3,982), \\ \texttt{bicycle}(2,412), \texttt{construction\_barrel}(2,899),\\ \texttt{wheeled\_device}(1,139), \texttt{bollard}(612),  \\ \texttt{mobile\_pedestrian\_crossing\_sign}(65)\end{tabular}}   
\\
\bottomrule
\end{tabular}}
\label{tab:dataset_descript}
\end{table*}
\begin{figure}[ht]
    \begin{center}
    \includegraphics[width=\linewidth]{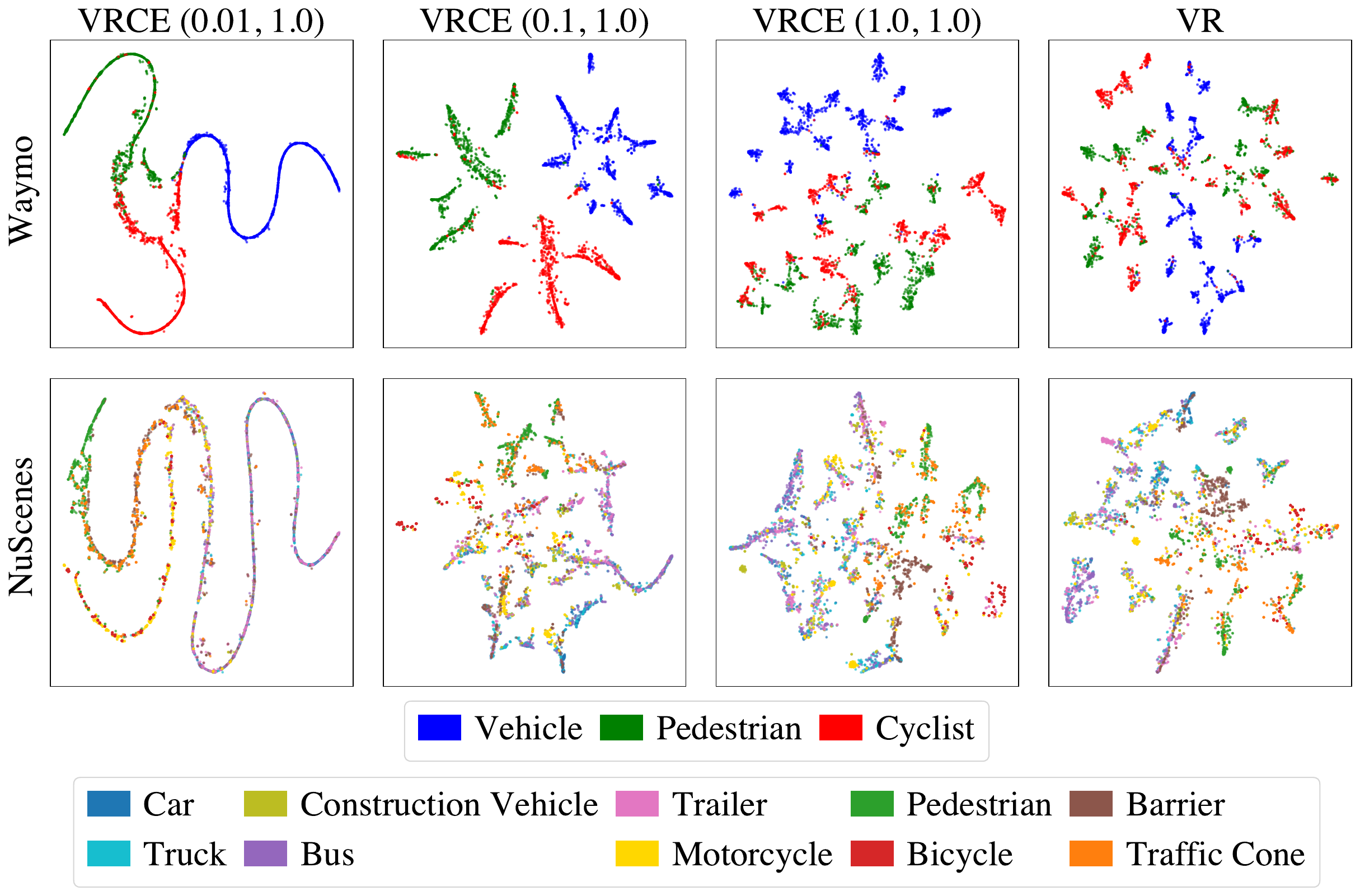}
   \caption{\textbf{t-SNE plots} of nuScenes samples, using the feature space of VRCE and VR pretrained models with different loss coefficients. From left to right, the strength of VR objective increases. The values in parentheses indicate the coefficients of the VR and CE objectives, respectively.}
    \label{fig:tsne_ssl_abl}
    \end{center}
\end{figure}

\subsection{Hybrid Pretraining Objectives}
NCECE and VRCE combine a self-supervised loss and a cross-entropy (CE) loss additively. 
To study how the self-supervised loss weight (SSL coefficient) affects knowledge transfer, we conduct an ablation on W$\rightarrow$N with the PointNet++ backbone. 
We fix the CE loss weight to 1.0 and vary the SSL coefficient in \{0.01, 0.1, 1.0\}. 
All experimental settings, model selection, and evaluation protocols follow those in the main paper.

\cref{tab:ssl_abl} reports the results. 
Both very small (0.01) and large (1.0) SSL coefficients degrade performance, whereas a moderate value (0.1) yields the best average performance across classes. 
Consistently, \cref{fig:tsne_ssl_abl} shows that a moderate SSL term encourages class embeddings to form coherent structures around their prototypes. 
By contrast, an excessively large SSL coefficient leads to cross-class mixing, while an overly small coefficient fails to mitigate CE-driven feature collapse.

We use an SSL coefficient of 0.1 as the default in the main paper; it gives the best ACC in both CL settings. 
Nevertheless, hybrid pretraining still does not outperform supervised pretraining under CL, as discussed in Section~4.4.

\subsection{$\lambda$ in Continual Learning}
\begin{figure}[t]
    \begin{center}
    \includegraphics[width=\columnwidth]{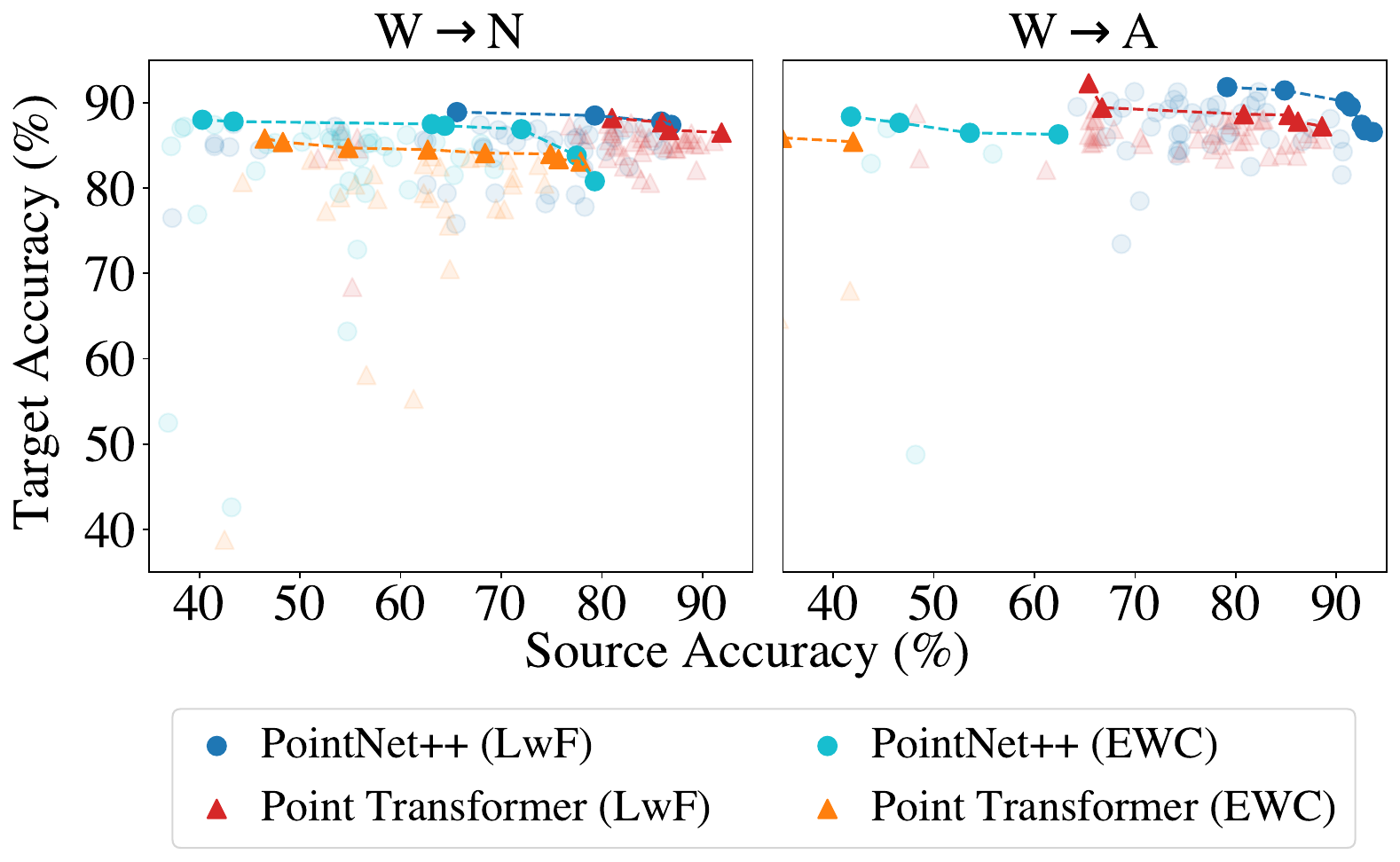}
   \caption{\textbf{Continual learning performance} on W$\rightarrow$N and W$\rightarrow$A under different design choices, shown separately for each CL method–backbone pair. Dashed lines denote the Pareto frontier.}
   \label{fig:pareto_summary}
   \end{center}
\end{figure}

We sweep the regularization weight $\lambda$ over \{0.25, 0.5, 0.75, 1.0, 1.25, 1.5\} for LwF and \{0.0005, 0.001, 0.005, 0.01, 0.05, 0.1\} for EWC, and select the best model based on ACC. 
Varying $\lambda$ reveals a trade-off between source and target performance, forming a Pareto frontier (\cref{fig:pareto_summary}).

EWC is more sensitive to $\lambda$, showing larger accuracy variance and consistently lower performance than LwF. 
This behavior aligns with prior observations \citep{DBLP:journals/corr/abs-2109-10021}, which report disproportionately large importance scores for neurons in convolutional and attention layers—core components of PointNet++ and Point Transformer—relative to linear layers, leading to optimization instability. 
Mitigating these stability issues for such backbones likely requires additional stabilization strategies, which we leave for future work.

\subsection{Domain Gap Reduction via Pre-Processing}
We employ \textbf{line downsampling} \citep{wei2022lidardistillationbridgingbeaminduced} and \textbf{bounding box normalization} to mitigate domain gaps that are not addressed in typical 3D object detection pipelines \citep{openpcdet2020}.
We ablate these factors in \cref{tab:whl_abl}.
Line downsampling improves ZS performance, despite a slight reduction in Waymo validation accuracy.
Bounding box normalization, however, exhibits dataset-dependent effects.
While differences in object dimensions $(l, w, h)$ provide useful cues for classification in a static setting, they can bias the model toward the source dataset and hinder adaptation to new domains.
Consistent with this intuition, bounding box normalization improves performance in W$\rightarrow$N transfer but slightly degrades performance in W$\rightarrow$A.

\begin{table}[h]
\centering
\caption{\textbf{Accuracy (\%)} in the ablation study on data preprocessing. \textbf{LD} and \textbf{BBN} denote line downsampling and bounding box normalization, respectively. The \textbf{Waymo} column reports $R_{S,f_S}$; the \textbf{W$\rightarrow$N} and \textbf{W$\rightarrow$A} columns report $R_{T,f_S}$. Best models in each column are highlighted in \textbf{bold}.}
\begin{tabular}{ccccc}
    \toprule
    \textbf{LD} &\textbf{BBN} & \textbf{Waymo} & \textbf{W$\rightarrow$N} & \textbf{W$\rightarrow$A} \\
    \midrule
    \midrule
    \cmark & \cmark & 95.6 & \textbf{89.8} & 75.7 \\
    \xmark & \cmark & \textbf{96.7} & 85.9 & 74.3 \\
    \cmark & \xmark & 96.1 & 89.2 & \textbf{79.0} \\
    \bottomrule
\end{tabular}
\label{tab:whl_abl}
\vskip -0.1in
\end{table}

\section{Embedding Space Visualization}
For every t-SNE plot in main body and appendix, We use the Scikit-Learn implementation \citep{scikit-learn} with perplexity set to 50 and all other hyperparameters kept at their default values. 

\begin{figure*}[t]
   \begin{center}
   \includegraphics[width=\linewidth]{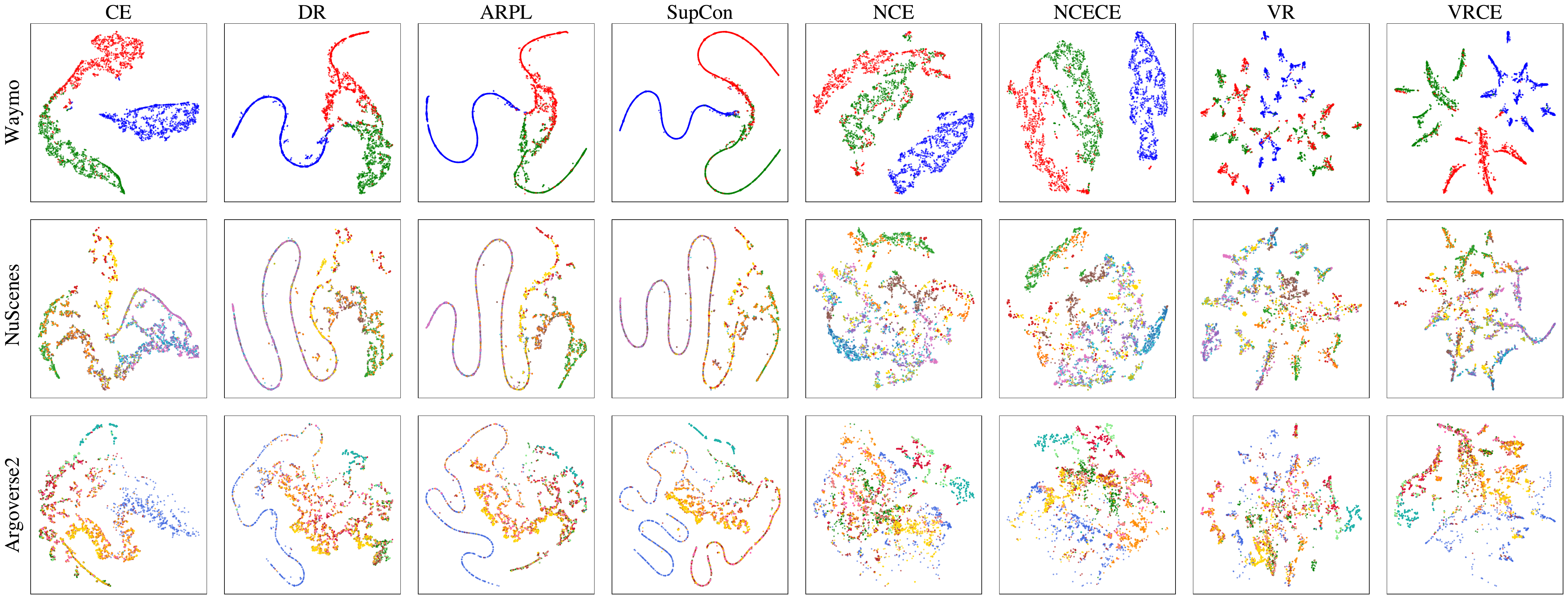} \\
    \includegraphics[width=\linewidth]{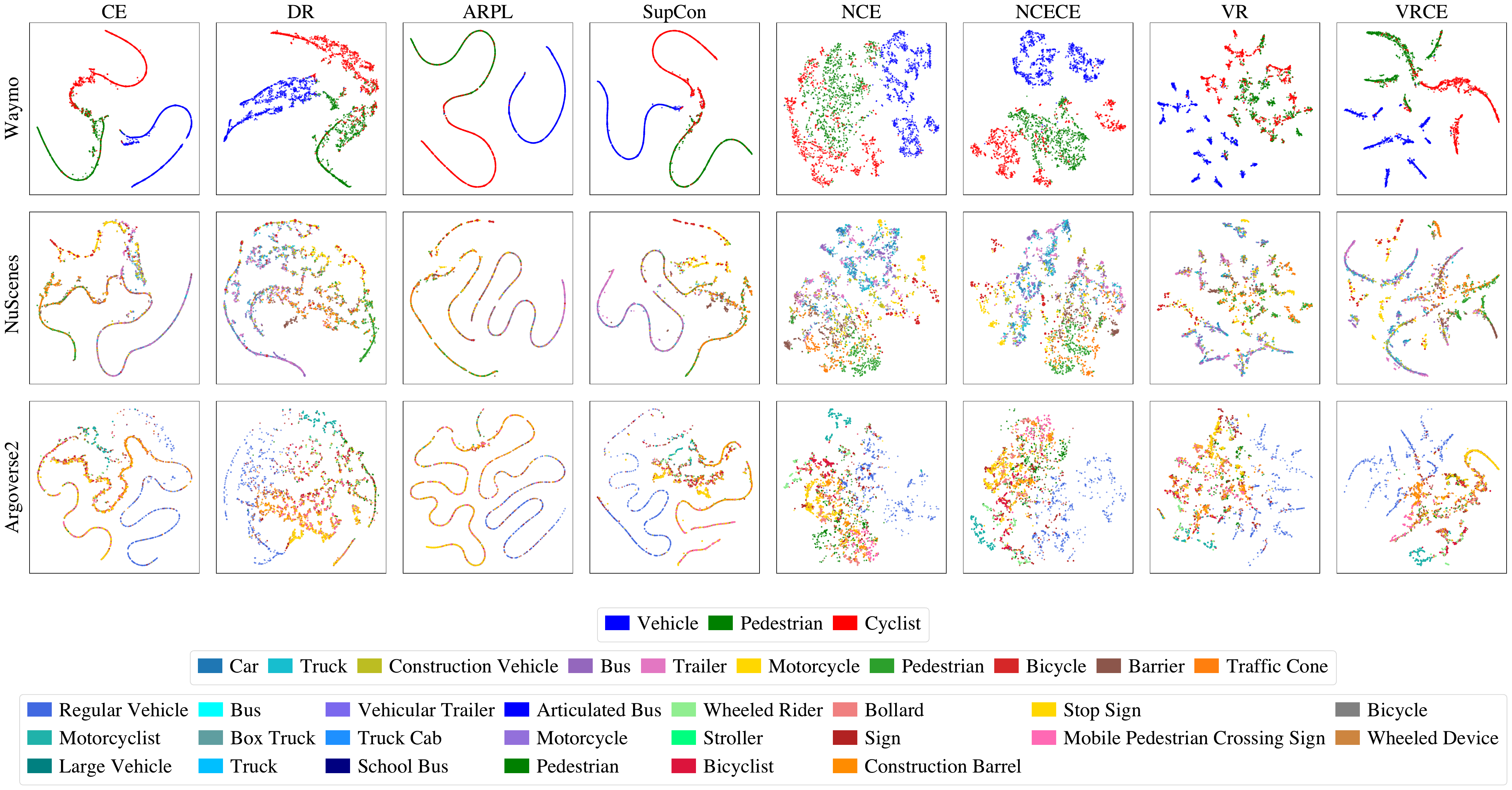}
   \caption{\textbf{t-SNE plots} of Waymo-pretrained models: Rows 1–3 show PointNet++ backbones, and Rows 4–6 show Point Transformer backbones.}
    \label{fig:tsne}
    \end{center}
\end{figure*}

In \cref{fig:tsne}, we visualize t-SNE plots of the penultimate feature space of Waymo-pretrained models for each pretraining objective. 
The pretrained feature geometry varies substantially across training objectives, which—consistent with Section 4.2—corresponds to large performance gaps in LP, particularly under label split and insertion.

\begin{figure}[t]
    \centering
    \includegraphics[width=\linewidth]{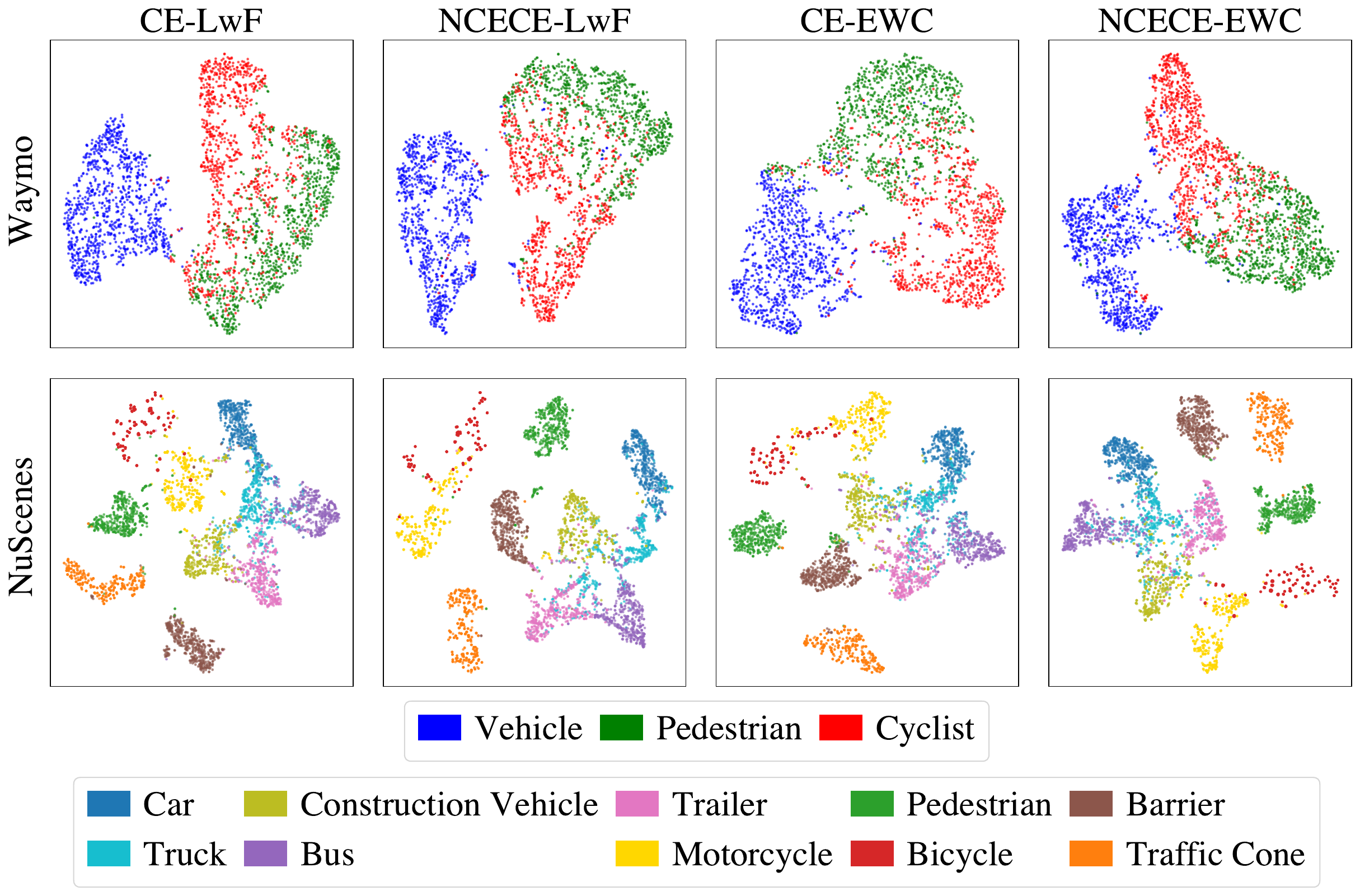}
   \caption{\textbf{t-SNE plots} of CE+PointNet++ and NCECE+PointNet++ models under W$\rightarrow$N continual learning. Columns 1–2 show LwF models, while Columns 3–4 show EWC models.}
    \label{fig:tsne_cl}
\end{figure}

After CL, however, these objective-dependent differences largely diminish (\cref{fig:tsne_cl}). 
This observation is consistent with Section 4.4: self-supervised pretraining no longer dominates supervised pretraining in CL, and the choice of the pretraining objective has a weaker effect in more flexible transfer settings where the representation is allowed to evolve.

\section{Continual Learning with Memory: LwF with Extra-Memory}
In the main paper, we assumed a general CL setting where no external data beyond the target stream is available during training. 
In practice, however, retaining a small buffer of samples from previous tasks (i.e., the source dataset) can substantially mitigate catastrophic forgetting \citep{DBLP:conf/nips/RolnickASLW19, lopez2017gradient, DBLP:phd/hal/Lesort20}, improving both BWT and ACC. 

Here, we evaluate W$\rightarrow$N LwF setting with additional memory across different pretraining objectives, using a PointNet++ backbone. 
During CL, the cross-entropy loss for the source head is computed on both the source memory and target training samples while knowledge is distilled from the pretrained teacher to the fine-tuned student. 
We use a random buffer of 3,000 source samples (1,000 per class in $\mathcal{C}_S$) and report average performance over five seeds.

As shown in \cref{tab:nusc_lwf_memory}, memory improves BWT for all objectives except DR. 
DR still suffers from the feature collapse problem mentioned in Section 4.3.
CE, SupCon, NCE, and NCECE also yield higher ACC. Notably, NCECE exhibits the largest gains with memory.

\begin{table}[hb]
\centering
\caption{\textbf{Accuracy (ACC and BWT) (\%)} of W$\rightarrow$N LwF continual learning models with additional Waymo memory. Values in parentheses indicate the standard deviation over five random seeds. \textbf{$\Delta$ACC} and \textbf{$\Delta$BWT} denote performance gains relative to models without additional memory. Best results in each column are highlighted in \textbf{bold}.}
\resizebox{\linewidth}{!}{
\begin{tabular}{lcccc}
\toprule
\multicolumn{1}{c}{}
& \textbf{ACC} & \textbf{BWT} & \textbf{$\Delta$ACC} & \textbf{$\Delta$BWT} \\
\midrule
\midrule
\textbf{CE} &88.7 \small{($\pm$1.0)} &-2.2 \small{($\pm$1.0)} &+6.2 &+15.4 \\
\textbf{DR} &61.3 \small{($\pm$1.2)} &-59.6 \small{($\pm$1.2)} &-9.0 &-14.3 \\
\textbf{ARPL} &85.4 \small{($\pm$2.7)} &-4.0 \small{($\pm$2.6)} &-1.5 &+5.7 \\
\textbf{SupCon} &88.2 \small{($\pm$0.5)} &-3.0 \small{($\pm$1.9)} &+1.8 &+6.9 \\
\textbf{NCE} &88.4 \small{($\pm$2.7)} &\textbf{-0.1} \small{($\pm$0.8)} &+6.0 &+13.3 \\
\textbf{NCECE} &\textbf{90.6} \small{($\pm$0.5)} &-3.1 \small{($\pm$0.6)} &\textbf{+6.6} &\textbf{+15.9} \\
\textbf{VR} &84.3 \small{($\pm$1.9)} &-3.4 \small{($\pm$3.1)} &-0.2 &+2.6 \\
\textbf{VRCE} &89.1 \small{($\pm$0.9)} &-2.8 \small{($\pm$1.2)} &-8.1 &+7.5 \\
\bottomrule
\end{tabular}}
\label{tab:nusc_lwf_memory}
\vskip -0.1in
\end{table}

\section{Additional Tables}\label{appendix-sec:full_table}
Full tables for W$\rightarrow$N and W$\rightarrow$A are provided: zero-shot (ZS) in \cref{tab:zeroshot}, linear probing (LP) in \cref{tab:nusc_lp,tab:argo2_lp}, and continual learning (CL) in \cref{tab:nusc_lwf,tab:nusc_ewc,tab:argo2_lwf,tab:argo2_ewc}.

\begin{table*}[t]
\centering
\caption{\textbf{Class-wise accuracy (\%)} of source dataset validation and zero-shot transfer (\textbf{ZS}) from W$\rightarrow$N and W$\rightarrow$A. 
The \textbf{Waymo} columns report $R_{S,f_S}^c$; the \textbf{W$\rightarrow$N} and \textbf{W$\rightarrow$A} columns report $R_{T,f_S}^c$.
\textbf{Veh.}, \textbf{Ped.}, and \textbf{Cyc.} denote the \texttt{vehicle}, \texttt{pedestrian}, and \texttt{cyclist}, respectively. 
For target dataset accuracies in zero-shot transfer, class names are followed by the type of label shift: \textbf{S} = split, \textbf{E} = expansion, \textbf{SE} = split + expansion, \textbf{M} = maintained.
Best models within each column and backbone are highlighted in \textbf{bold}.}
\resizebox{\linewidth}{!}{
\begin{tabularx}{\linewidth}{llYYY|YYY|YYYY}
\toprule
&                 & \multicolumn{3}{c}{\textbf{Waymo}}                                                    & \multicolumn{3}{c}{\textbf{W$\rightarrow$N (ZS)}} & \multicolumn{4}{c}{\textbf{W$\rightarrow$A (ZS)}}                                                     \\
\cmidrule(lr){3-5}\cmidrule(lr){6-8}\cmidrule(lr){9-12}
&                 & \textbf{Veh.}             & \textbf{Ped.}          & \multicolumn{1}{c}{\textbf{Cyc.}}             & \textbf{Veh.S.}            & \textbf{Ped.M.}           & \multicolumn{1}{c}{\textbf{Cyc.E.}}             &
\textbf{Veh.S.}            & \textbf{Ped.S.}          &
\textbf{Ped.SE.}             & \textbf{Cyc.M.} 
\\
\midrule
\midrule
\multicolumn{1}{l|}{}                                      & \textbf{CE}     & 98.5                                     & {\textbf{99.4}} & 88.8 & 91.7 & {\textbf{98.2}} & {\textbf{70.2}} & 77.1 & {\textbf{97.1}} & 91.6 & 1.6 \\
\multicolumn{1}{l|}{}                                      & \textbf{DR}    & {\textbf{99.7}}                                     & 96.8 & {\textbf{96.8}} & {\textbf{99.7}} & 85.9 & 17.5 & {\textbf{87.3}} & 79.0 & 69.0 & 1.9 \\
\multicolumn{1}{l|}{}                                      & \textbf{ARPL} & 99.3                                     & 96.9 & 89.1 & 95.7 & 85.8 & 64.9 & 81.9 & 87.0 & 38.0 & {\textbf{5.9}} \\
\multicolumn{1}{l|}{}                                      & \textbf{SupCon} & 98.4                                     & 96.1 & 92.7 & 93.7 & 83.4 & 49.1 & 79.7 & 86.8 & 77.5 & 4.4 \\
\cmidrule{2-12}
\multicolumn{1}{l|}{}                                      & \textbf{NCE}   & 99.3                                     & 93.2 & 86.5 & 95.1 & 61.4 & 52.6 & 77.4 & 91.3 & 19.7 & 4.4 \\
\multicolumn{1}{l|}{}                                      & 
\textbf{NCECE}  & 99.5                                     & 98.8 & 95.3 & 96.3 & 95.3 & 59.7 & 82.7 & 95.4 & {\textbf{95.8}} & 1.6 \\
\multicolumn{1}{l|}{}                                      & \textbf{VR}    & 96.6                                     & 85.0   & 80.7 & 97.8 & 20.2 & 17.5 & 59.3 & 92.9 & 80.3 & 2.8 \\
\multicolumn{1}{l|}{\multirow{-8}{*}{\rotatebox{90}{\textbf{PointNet++}}}} & \textbf{VRCE}   & 99.7                                     & 97.0   & 94.0   & 96.5 &82.1 & 52.6 & 85.6 & 90.1 & 85.9 &5.0
\\
\midrule
\multicolumn{1}{l|}{}                                      & \textbf{CE}     & 99.4                                     & 98.1 & {\textbf{92.1}} & {\textbf{93.5}}& {\textbf{96.4}} & 54.4 & 85.1 & 93.6 & 95.8 & 1.2 \\
\multicolumn{1}{l|}{}                                      & \textbf{DR}     & {\textbf{99.5}}                                     & 91.9 & 83.8 & {\textbf{93.5}} & 89.9 & 24.6 & 90.5 & 67.5 & 90.1 & {\textbf{5.9}} \\
\multicolumn{1}{l|}{}                                      & \textbf{ARPL}   & 98.7                                     & 97.1 & 89.8 &92.2 & 92.9 & 59.7 & 86.3 & 84.1 & {\textbf{97.2}}& 2.5 \\
\multicolumn{1}{l|}{}                                      & \textbf{SupCon} & 98.9                                     & 97.1 & 91.3 & 91.6 & 90.4 & 68.4 & 88.0 & 80.5 & 95.8 & 1.2 \\
\cmidrule{2-12}
\multicolumn{1}{l|}{}
& \textbf{NCE}    & 98.4                                     & 88.0 &89.4 & 89.6 & 82.5 & 63.2 & 87.4 & 60.4 & 69.0 & 0.6 \\
\multicolumn{1}{l|}{}                                      & \textbf{NCECE}  & 99.0                                     & 98.6 & 91.0 & 89.4 & 92.8 & 50.9 & 84.7 & 86.2 & 85.9 & 1.2 \\
\multicolumn{1}{l|}{}                                      & \textbf{VR}     & 98.2                                     & 86.7 & 79.0 & 87.0 & 74.8 & 75.4 & {\textbf{92.2}} & 51.1 & 74.6 & 0.9 \\
\multicolumn{1}{l|}{\multirow{-8}{*}{\rotatebox{90}{\textbf{Point Transformer} }}} & \textbf{VRCE}   & 98.9                                     & {\textbf{99.1}} & 87.6 & 91.3 & 95.9 & {\textbf{79.0}} & 85.8 & {\textbf{99.1}} & 95.8 & 0.6
\\
\bottomrule
\end{tabularx}}
\label{tab:zeroshot}
\end{table*}
\begin{table}[b]
\centering
\caption{\textbf{Class-wise accuracy (\%)} of linear probing (\textbf{LP}) from W$\rightarrow$N. 
Each column reports $R_{T,f_T}^c$.
\textbf{Veh.}, \textbf{Ped.}, \textbf{Cyc.}, and \textbf{Insert.} denote the \texttt{vehicle}, \texttt{pedestrian}, \texttt{cyclist}, and inserted classes, respectively. 
Class names are followed by the type of label shift: \textbf{S} = split, \textbf{E} = expansion, and \textbf{M} = maintained.
Best models within each column and backbone are highlighted in \textbf{bold}.}
\resizebox{\linewidth}{!}{
\begin{tabularx}{\linewidth}{ll*{4}{Y}}
\toprule
\multicolumn{2}{l}{\textbf{W$\rightarrow$N (LP)}} & \textbf{Veh.S.} & \textbf{Ped.M.} & \textbf{Cyc.E.} & \textbf{Insert.} \\
\midrule
\midrule
\multicolumn{1}{l|}{} & \textbf{CE}     & 21.4 & 88.0 & 54.4 & 38.5 \\
\multicolumn{1}{l|}{} & \textbf{DR}     & 23.4 & 82.4 & {\textbf{93.0}} & 40.1 \\
\multicolumn{1}{l|}{} & \textbf{ARPL}   & 20.7 & {\textbf{90.3}} & 73.7 & 26.3 \\
\multicolumn{1}{l|}{} & \textbf{SupCon} & 19.9 & 84.0 & 38.6 & 18.2 \\
\cmidrule{2-6}
\multicolumn{1}{l|}{} & \textbf{NCE}    & 40.4 & 80.3 & 40.4 & 71.5 \\
\multicolumn{1}{l|}{} & \textbf{NCECE}  & {\textbf{44.6}} & 82.6 & 84.2 & 70.7 \\
\multicolumn{1}{l|}{} & \textbf{VR}     & 32.4 & 72.3 & 61.4 & {\textbf{72.5}} \\
\multicolumn{1}{l|}{\multirow{-8}{*}{\rotatebox{90}{\textbf{PointNet++}}}} & \textbf{VRCE}   & 32.3 & 52.5 & 80.7 & 57.1 \\
\midrule
\multicolumn{1}{l|}{} & \textbf{CE}     & 19.7 & 89.1 & 86.0 & 16.8 \\
\multicolumn{1}{l|}{} & \textbf{DR}     & 24.2 & {\textbf{90.0}} & 82.5 & 20.8 \\
\multicolumn{1}{l|}{} & \textbf{ARPL}   & 16.0 & 83.7 & 91.2 & 15.6 \\
\multicolumn{1}{l|}{} & \textbf{SupCon} & 21.0 & 80.3 & {\textbf{98.3}} & 17.9 \\
\cmidrule{2-6}
\multicolumn{1}{l|}{} & \textbf{NCE}    & {\textbf{44.4}} & 77.0 & 66.7 & {\textbf{57.7}} \\
\multicolumn{1}{l|}{} & \textbf{NCECE}  & 44.2 & 79.1 & 86.0 & 50.8 \\
\multicolumn{1}{l|}{} & \textbf{VR}     & 34.6 & 70.4 & 63.2 & 48.4 \\
\multicolumn{1}{l|}{\multirow{-8}{*}{\rotatebox{90}{\textbf{Point Transformer}}}} & \textbf{VRCE} & 33.9 & 78.9 & 80.7 & 45.7 \\
\bottomrule
\end{tabularx}}
\label{tab:nusc_lp}
\end{table}

\begin{table}[b]
\centering
\caption{\textbf{Class-wise accuracy (\%)} of linear probing (\textbf{LP}) from W$\rightarrow$A. 
Each column reports $R_{T,f_T}^c$.
\textbf{Veh.}, \textbf{Ped.}, \textbf{Cyc.}, and \textbf{Insert.} denote the \texttt{vehicle}, \texttt{pedestrian}, \texttt{cyclist}, and inserted classes, respectively. 
Class names are followed by the type of label shift: \textbf{S} = split, \textbf{E} = expansion, \textbf{SE} = split + expansion, and \textbf{M} = maintained.
Best models within each column and backbone are highlighted in \textbf{bold}.}
\resizebox{\linewidth}{!}{
\begin{tabularx}{\linewidth}{ll*{5}{Y}}
\toprule
\multicolumn{2}{l}{\textbf{W$\rightarrow$A LP}}          & \textbf{Veh.S.} & \textbf{Ped.S.}          & \textbf{Ped.SE.}           & \textbf{Cyc.M}             & \textbf{Insert.}          \\
\midrule
\midrule
\multicolumn{1}{l|}{}                                      & \textbf{CE}     & 9.4  & 39.5 & 42.2 & 24.6 & 23.0 \\
\multicolumn{1}{l|}{}                                      & \textbf{DR}     & 8.8  & 35.6 & 42.2 & 10.9 & 20.8 \\
\multicolumn{1}{l|}{}                                      & \textbf{ARPL}   & 10.3 & 25.5 & 2.8  & 24.9 & 23.6 \\
\multicolumn{1}{l|}{}                                      & \textbf{SupCon} & 9.0  & 12.7 & 21.1 & 5.0  & 23.7 \\
\cmidrule{2-7}
\multicolumn{1}{l|}{}                                      & \textbf{NCE}    & 25.1 & 48.0 & \textbf{83.1} & 48.3 & 36.5 \\
\multicolumn{1}{l|}{}                                      & \textbf{NCECE}  & \textbf{27.1} & \textbf{50.3} & 80.3 & \textbf{49.5} & \textbf{44.4} \\
\multicolumn{1}{l|}{}                                      & \textbf{VR}     & 14.7 & 24.1 & 32.4 & 48.3 & 35.9 \\
\multicolumn{1}{l|}{\multirow{-8}{*}{\rotatebox{90}{\textbf{PointNet++}}}} & \textbf{VRCE}   & 15.9 & 55.9 & 52.1 & 19.3 & 32.1 \\
\midrule
\multicolumn{1}{l|}{}                                      & \textbf{CE}     & 9.8  & \textbf{34.2} & 0.0  & 5.9  & 12.6 \\
\multicolumn{1}{l|}{}                                      & \textbf{DR}     & 9.4  & 31.1 & 0.0  & 15.9 & 21.7 \\
\multicolumn{1}{l|}{}                                      & \textbf{ARPL}   & 8.4  & 0.0  & 1.4  & 0.0  & 24.4 \\
\multicolumn{1}{l|}{}                                      & \textbf{SupCon} & 9.1  & 0.0  & 9.9  & 0.0  & 19.4 \\
\cmidrule{2-7}
\multicolumn{1}{l|}{}                                      & \textbf{NCE}    & 26.4 & 18.5 & 35.2 & 40.5 & \textbf{40.0} \\
\multicolumn{1}{l|}{}                                      & \textbf{NCECE}  & \textbf{27.3} & 22.4 & 25.4 & 31.8 & 39.9 \\
\multicolumn{1}{l|}{}                                      & \textbf{VR}     & 15.2 & 18.4 & \textbf{38.0} & \textbf{57.9} & 28.4 \\
\multicolumn{1}{l|}{\multirow{-8}{*}{\rotatebox{90}{\textbf{Point Transformer}}}} & \textbf{VRCE}   & 14.4 & 21.5 & 36.6 & 39.2 & 32.3 \\
\bottomrule
\end{tabularx}}
\label{tab:argo2_lp}
\end{table}
\begin{table*}[t]
\centering
\caption{\textbf{ACC, BWT, and class-wise accuracy (\%)} of continual learning with \textbf{LwF} from W$\rightarrow$N. 
The \textbf{nuScenes} columns report $R_{T,f_T}^c$; the \textbf{Waymo} columns report $R_{S,f_T}^c$.
\textbf{Veh.}, \textbf{Ped.}, \textbf{Cyc.}, and \textbf{Insert.} denote the \texttt{vehicle}, \texttt{pedestrian}, \texttt{cyclist}, and inserted classes respectively. 
Class names are followed by the type of label shift: \textbf{S} = split, \textbf{E} = expansion, and \textbf{M} = maintained.
Best models within each column and backbone are highlighted in \textbf{bold}.}
\resizebox{\linewidth}{!}{
\begin{tabularx}{\linewidth}{llYY|YYYY|YYY}
\toprule
\multicolumn{2}{l}{} & \multicolumn{1}{c}{} &\multicolumn{1}{c}{}  & \multicolumn{4}{c}{\textbf{nuScenes}} & \multicolumn{3}{c}{\textbf{Waymo}} \\
\cmidrule(lr){5-8}\cmidrule(lr){9-11}
\multicolumn{2}{l}{\multirow{-2}{*}{\textbf{W$\rightarrow$N (LwF)}}}
& \multicolumn{1}{c}{\multirow{-2}{*}{\textbf{ACC}}}
& \multicolumn{1}{c}{\multirow{-2}{*}{\textbf{BWT}}}
& \multicolumn{1}{c}{\textbf{Veh.S.}}
& \multicolumn{1}{c}{\textbf{Ped.M.}}
& \multicolumn{1}{c}{\textbf{Cyc.E.}}
& \multicolumn{1}{c}{\textbf{Insert.}}
& \textbf{Veh.}
& \textbf{Ped.}
& \textbf{Cyc.} \\
\midrule
\midrule
\multicolumn{1}{l|}{} & \textbf{CE}     & 82.5 & -17.6 & 82.7 & 79.0 & 94.7 & 96.1 & 99.4 & {\textbf{98.7}} & 38.1 \\
\multicolumn{1}{l|}{} & \textbf{DR}     & 70.3 & -45.3 & 82.5 & 90.5 & 89.5 & 98.6 & 27.0 & 97.8 & 35.6 \\
\multicolumn{1}{l|}{} & \textbf{ARPL}   & 86.9 & -9.7  & 82.7 & 91.8 & 93.0 & 98.9 & 97.8 & 84.4 & 75.5 \\
\multicolumn{1}{l|}{} & \textbf{SupCon} & 86.4 & -9.9  & 80.5 & 94.0 & 94.7 & 97.2 & 99.0 & 91.2 & 68.4 \\
\cmidrule{2-11}
\multicolumn{1}{l|}{} & \textbf{NCE}    & 82.4 & -13.4 & 78.4 & 92.8 & 82.5 & {\textbf{99.2}} & 99.0 & 78.2 & 64.2 \\
\multicolumn{1}{l|}{} & \textbf{NCECE}  & 84.0 & -19.0 & {\textbf{83.3}} & 94.0 & {\textbf{96.5}} & 98.5 & {\textbf{99.5}} & 96.5 & 41.9 \\
\multicolumn{1}{l|}{} & \textbf{VR}     & 84.5 & \textbf{-6.1}  & 81.3 & {\textbf{98.9}} & 89.5 & 96.3 & 90.2 & 74.7 & {\textbf{81.6}} \\
\multicolumn{1}{l|}{\multirow{-8}{*}{\rotatebox{90}{\textbf{PointNet++}}}}
& \textbf{VRCE} & {\textbf{87.2}} & -10.3 & 81.7 & 95.6 & 91.2 & 98.6 & {\textbf{99.5}} & 93.6 & 67.5 \\
\midrule
\multicolumn{1}{l|}{} & \textbf{CE}     & {\textbf{89.2}} & -4.8  & 80.5 & 90.6 & 96.5 & 97.7 & 97.3 & 96.7 & 81.7 \\
\multicolumn{1}{l|}{} & \textbf{DR}     & 70.8 & -39.2 & 80.2 & 93.6 & 91.2 & 96.8 & 2.9  & 88.6 & 75.6 \\
\multicolumn{1}{l|}{} & \textbf{ARPL}   & 84.9 & -11.9 & 79.6 & 93.4 & 93.0 & 97.0 & 93.8 & 88.8 & 69.2 \\
\multicolumn{1}{l|}{} & \textbf{SupCon} & 87.2 & -6.9  & 78.4 & {\textbf{93.7}} & 96.5 & 95.6 & 96.1 & 95.7 & 75.8 \\
\cmidrule{2-11}
\multicolumn{1}{l|}{} & \textbf{NCE}    & 87.7 & \textbf{-3.3}  & {\textbf{80.9}} & 87.1 & 96.5 & 97.9 & {\textbf{99.1}} & 76.5 & {\textbf{91.2}} \\
\multicolumn{1}{l|}{} & \textbf{NCECE}  & 87.4 & -6.7  & 78.2 & 89.3 & {\textbf{98.3}} & 97.4 & 98.1 & 95.4 & 75.8 \\
\multicolumn{1}{l|}{} & \textbf{VR}     & 85.3 & -3.8  & 80.0 & 89.8 & 93.0 & {\textbf{98.7}} & 95.6 & 92.6 & 65.9 \\
\multicolumn{1}{l|}{\multirow{-8}{*}{\rotatebox{90}{\textbf{Point Transformer}}}}
& \textbf{VRCE} & 86.0 & -8.2 & 79.1 & 91.3 & 87.7 & 96.5 & 97.8 & {\textbf{97.4}} & 66.9 \\
\bottomrule
\end{tabularx}}
\label{tab:nusc_lwf}
\end{table*}

\begin{table*}[b]
\centering
\caption{\textbf{ACC, BWT, and class-wise accuracy (\%)} of continual learning with \textbf{EWC} from W$\rightarrow$N. 
The \textbf{nuScenes} columns report $R_{T,f_T}^c$; the \textbf{Waymo} columns report $R_{S,f_T}^c$.
\textbf{Veh.}, \textbf{Ped.}, \textbf{Cyc.}, and \textbf{Insert.} denote the \texttt{vehicle}, \texttt{pedestrian}, \texttt{cyclist}, and inserted classes respectively. 
Class names are followed by the type of label shift: \textbf{S} = split, \textbf{E} = expansion, and \textbf{M} = maintained.
Best models within each column and backbone are highlighted in \textbf{bold}.}
\resizebox{\linewidth}{!}{
\begin{tabularx}{\linewidth}{llYY|YYYY|YYY}
\toprule
\multicolumn{2}{l}{} &\multicolumn{1}{c}{}  &\multicolumn{1}{c}{}  & \multicolumn{4}{c}{\textbf{nuScenes}} & \multicolumn{3}{c}{\textbf{Waymo}} \\
\cmidrule(lr){5-8}\cmidrule(lr){9-11}
\multicolumn{2}{l}{\multirow{-2}{*}{\textbf{W$\rightarrow$N (EWC)}}}
& \multicolumn{1}{c}{\multirow{-2}{*}{\textbf{ACC}}}
& \multicolumn{1}{c}{\multirow{-2}{*}{\textbf{BWT}}}
& \textbf{Veh.S.} & \textbf{Ped.M.} & \textbf{Cyc.E.} & \multicolumn{1}{c}{\textbf{Insert.}}
& \textbf{Veh.} & \textbf{Ped.} & \textbf{Cyc.} \\
\midrule
\midrule
\multicolumn{1}{l|}{} & \textbf{CE}     & 74.8 & -31.1 & 76.8 & 96.3 & 84.2 & 97.7 & 47.9 & \textbf{95.7} & 54.1 \\
\multicolumn{1}{l|}{} & \textbf{DR}     & 79.5 & -26.3 & 80.6 & 96.1 & 93.0 & 98.1 & 66.0 & 74.7 & 75.4 \\
\multicolumn{1}{l|}{} & \textbf{ARPL}   & 75.8 & -32.3 & \textbf{81.4} & 96.6 & 93.0 & 97.5 & 65.0 & 78.0 & 50.1 \\
\multicolumn{1}{l|}{} & \textbf{SupCon} & 73.4 & -31.8 & 72.8 & 85.5 & 96.5 & \textbf{98.5} & 49.3 & 69.5 & 77.2 \\
\cmidrule{2-11}
\multicolumn{1}{l|}{} & \textbf{NCE}    & 73.4 & -34.8 & 81.0 & \textbf{96.8} & 84.2 & 97.8 & \textbf{76.1} & 76.2 & 29.5 \\
\multicolumn{1}{l|}{} & \textbf{NCECE}  & \textbf{80.7} & -20.7 & 74.9 & 93.4 & \textbf{98.3} & \textbf{98.5} & 69.2 & 82.6 & 80.9 \\
\multicolumn{1}{l|}{} & \textbf{VR}     & 71.0 & -35.2 & 78.3 & 95.7 & 91.2 & 97.9 & 67.2 & 76.2 & 26.6 \\
\multicolumn{1}{l|}{\multirow{-8}{*}{\rotatebox{90}{\textbf{PointNet++}}}}
& \textbf{VRCE}   & 80.1 & \textbf{-18.2} & 71.7 & 92.9 & 96.5 & 94.6 & 63.6 & 84.5 & \textbf{89.8} \\
\midrule
\multicolumn{1}{l|}{} & \textbf{CE}     & \textbf{80.5} & -19.3 & \textbf{81.6} & 80.8 & 80.7 & 90.4 & \textbf{86.3} & 69.5 & \textbf{77.9} \\
\multicolumn{1}{l|}{} & \textbf{DR}     & 79.5 & \textbf{-18.3} & 79.2 & 89.7 & 82.5 & 96.7 & 75.1 & 79.1 & 70.6 \\
\multicolumn{1}{l|}{} & \textbf{ARPL}   & 67.7 & -31.8 & 68.1 & 88.8 & 26.3 & 91.0 & 76.3 & \textbf{91.5} & 27.0 \\
\multicolumn{1}{l|}{} & \textbf{SupCon} & 77.5 & -22.5 & 74.6 & 89.9 & 77.2 & 95.7 & 72.0 & 81.2 & 69.7 \\
\cmidrule{2-11}
\multicolumn{1}{l|}{} & \textbf{NCE}    & 75.9 & -25.9 & 76.1 & \textbf{95.7} & \textbf{93.0} & 95.4 & 73.6 & 89.0 & 41.7 \\
\multicolumn{1}{l|}{} & \textbf{NCECE}  & 76.3 & -28.9 & 77.9 & 88.5 & 91.2 & \textbf{97.3} & 72.7 & 65.6 & 67.0 \\
\multicolumn{1}{l|}{} & \textbf{VR}     & 73.2 & -27.4 & 77.8 & 89.5 & 77.2 & 95.7 & 63.4 & 67.7 & 60.5 \\
\multicolumn{1}{l|}{\multirow{-8}{*}{\rotatebox{90}{\textbf{Point Transformer}}}}
& \textbf{VRCE}   & 73.9 & -26.2 & 67.7 & 92.8 & 87.7 & 94.6 & 65.8 & 71.4 & 73.7 \\
\bottomrule
\end{tabularx}}
\label{tab:nusc_ewc}
\end{table*}

\begin{table*}[t]
\centering
\caption{\textbf{ACC, BWT, and class-wise accuracy (\%)} of continual learning with \textbf{LwF} from W$\rightarrow$A. 
The \textbf{Argoverse2} columns report $R_{T,f_T}^c$; the \textbf{Waymo} columns report $R_{S,f_T}^c$.
\textbf{Veh.}, \textbf{Ped.}, \textbf{Cyc.}, and \textbf{Insert.} denote the \texttt{vehicle}, \texttt{pedestrian}, \texttt{cyclist}, and inserted classes respectively. 
Class names are followed by the type of label shift: \textbf{S} = split, \textbf{E} = expansion, \textbf{SE} = split+expansion, and \textbf{M} = maintained.
Best models within each column and backbone are highlighted in \textbf{bold}.}
\resizebox{\linewidth}{!}{
\begin{tabularx}{\linewidth}{llYY|YYYYY|YYY}
\toprule
\multicolumn{2}{l}{} & \multicolumn{1}{c}{} & \multicolumn{1}{c}{}  & \multicolumn{5}{c}{\textbf{Argoverse2}} & \multicolumn{3}{c}{\textbf{Waymo}} \\
\cmidrule(lr){5-9}\cmidrule(lr){10-12}
\multicolumn{2}{l}{\multirow{-2}{*}{\textbf{W$\rightarrow$A (LwF)}}}
& \multicolumn{1}{c}{\multirow{-2}{*}{\textbf{ACC}}}
& \multicolumn{1}{c}{\multirow{-2}{*}{\textbf{BWT}}}
& \textbf{Veh.S.} & \textbf{Ped.S.} & \textbf{Ped.SE.} & \textbf{Cyc.M.} & \multicolumn{1}{c}{\textbf{Insert.}}
& \textbf{Veh.} & \textbf{Ped.} & \textbf{Cyc.} \\
\midrule
\midrule
\multicolumn{1}{l|}{} & \textbf{CE}     & 73.6 & -13.1 & 53.8 & 66.8 & 67.6 & \textbf{69.5} & 78.4 & \textbf{99.9} & 87.6 & 61.5 \\
\multicolumn{1}{l|}{} & \textbf{DR}     & 71.7 & -20.2 & 56.0 & 65.2 & 83.1 & 64.5 & 77.3 & 71.1 & 91.3 & 71.9 \\
\multicolumn{1}{l|}{} & \textbf{ARPL}   & \textbf{78.3} & -4.4 & 54.9 & 78.4 & 97.2 & 38.0 & 78.4 & 98.5 & 80.3 & \textbf{93.9} \\
\multicolumn{1}{l|}{} & \textbf{SupCon} & 75.5 & -5.6 & 49.6 & 61.3 & 85.9 & 63.9 & 73.4 & 99.2 & 96.5 & 75.5 \\
\cmidrule{2-12}
\multicolumn{1}{l|}{} & \textbf{NCE}    & 69.9 & -20.5 & 55.0 & 79.1 & 62.0 & 55.1 & 81.4 & 99.8 & 94.8 & 27.0 \\
\multicolumn{1}{l|}{} & \textbf{NCECE}  & 73.4 & -19.1 & 54.9 & \textbf{80.3} & 97.2 & 53.0 & \textbf{81.9} & 99.5 & \textbf{99.2} & 38.8 \\
\multicolumn{1}{l|}{} & \textbf{VR}     & 76.0 & \textbf{-3.0} & 55.5 & 69.2 & \textbf{100.0} & 67.9 & 80.2 & 98.7 & 79.4 & 76.5 \\
\multicolumn{1}{l|}{\multirow{-8}{*}{\rotatebox{90}{\textbf{PointNet++}}}}
& \textbf{VRCE}   & 78.2 & -5.7 & \textbf{58.4} & 71.2 & 93.0 & 27.7 & 74.8 & 99.8 & 95.6 & 78.8 \\
\midrule
\multicolumn{1}{l|}{} & \textbf{CE}     & 74.1 & -9.3 & 49.2 & 44.8 & 91.6 & 81.9 & 75.5 & 99.3 & 98.7 & 64.7 \\
\multicolumn{1}{l|}{} & \textbf{DR}     & 71.4 & -13.2 & 57.6 & 60.2 & 26.8 & 74.1 & 76.6 & 85.6 & 96.2 & 57.0 \\
\multicolumn{1}{l|}{} & \textbf{ARPL}   & 66.8 & -31.3 & \textbf{58.7} & \textbf{81.5} & 95.8 & 55.4 & 77.0 & 97.6 & 98.6 & 0.0 \\
\multicolumn{1}{l|}{} & \textbf{SupCon} & 74.5 & -8.4 & 51.9 & 54.2 & 94.4 & 62.3 & 70.7 & 98.3 & 93.5 & 73.8 \\
\cmidrule{2-12}
\multicolumn{1}{l|}{} & \textbf{NCE}    & 74.3 & -5.1 & 50.8 & 41.4 & \textbf{97.2} & 73.8 & 76.6 & 97.6 & 85.7 & 78.4 \\
\multicolumn{1}{l|}{} & \textbf{NCECE}  & \textbf{75.4} & -7.9 & 56.8 & 37.1 & 91.6 & 67.0 & 73.3 & \textbf{99.6} & 81.6 & \textbf{84.5} \\
\multicolumn{1}{l|}{} & \textbf{VR}     & 74.6 & \textbf{-2.0} & 52.2 & 44.7 & 90.1 & \textbf{83.5} & \textbf{78.2} & 98.9 & 89.4 & 70.2 \\
\multicolumn{1}{l|}{\multirow{-8}{*}{\rotatebox{90}{\textbf{Point Transformer}}}}
& \textbf{VRCE}   & 70.7 & -12.5 & 50.5 & 56.2 & 40.9 & 48.6 & 74.5 & 99.1 & \textbf{99.2} & 51.7 \\
\bottomrule
\end{tabularx}}
\label{tab:argo2_lwf}
\end{table*}
\begin{table*}[b]
\centering
\caption{\textbf{ACC, BWT, and class-wise accuracy (\%)} of continual learning with \textbf{EWC} from W$\rightarrow$A. 
The \textbf{Argoverse2} columns report $R_{T,f_T}^c$; the \textbf{Waymo} columns report $R_{S,f_T}^c$.
\textbf{Veh.}, \textbf{Ped.}, \textbf{Cyc.}, and \textbf{Insert.} denote the \texttt{vehicle}, \texttt{pedestrian}, \texttt{cyclist}, and inserted classes respectively. 
Class names are followed by the type of label shift: \textbf{S} = split, \textbf{E} = expansion, \textbf{SE} = split+expansion, and \textbf{M} = maintained.
Best models within each column and backbone are highlighted in \textbf{bold}.}
\resizebox{\linewidth}{!}{
\begin{tabularx}{\linewidth}{llYY|YYYYY|YYY}
\toprule
\multicolumn{2}{l}{} & \multicolumn{1}{c}{} & \multicolumn{1}{c}{}  & \multicolumn{5}{c}{\textbf{Argoverse2}} & \multicolumn{3}{c}{\textbf{Waymo}} \\
\cmidrule(lr){5-9}\cmidrule(lr){10-12}
\multicolumn{2}{l}{\multirow{-2}{*}{\textbf{W$\rightarrow$A (EWC)}}}
& \multicolumn{1}{c}{\multirow{-2}{*}{\textbf{ACC}}}
& \multicolumn{1}{c}{\multirow{-2}{*}{\textbf{BWT}}}
& \textbf{Veh.S.} & \textbf{Ped.S.} & \textbf{Ped.SE.} & \textbf{Cyc.M.} & \multicolumn{1}{c}{\textbf{Insert.}}
& \textbf{Veh.} & \textbf{Ped.} & \textbf{Cyc.} \\
\midrule
\midrule
\multicolumn{1}{l|}{} & \textbf{CE}     & 52.7 & -56.3 & 51.1 & 70.2 & 66.2 & 48.9 & 83.2 & 17.0 & 39.7 & {\textbf{68.6}} \\
\multicolumn{1}{l|}{} & \textbf{DR}     & 57.5 & -45.2 & 50.3 & 66.3 & 64.8 & 45.2 & 79.3 & 67.4 & 93.3 & 0.0 \\
\multicolumn{1}{l|}{} & \textbf{ARPL}   & 39.6 & -84.2 & 51.8 & 61.9 & 85.9 & {\textbf{81.0}} & {\textbf{79.0}} & 17.0 & 1.1 & 26.9 \\
\multicolumn{1}{l|}{} & \textbf{SupCon} & 50.5 & -54.3 & 46.0 & 54.6 & 90.1 & 50.8 & 71.6 & 82.4 & 48.9 & 0.0 \\
\cmidrule{2-12}
\multicolumn{1}{l|}{} & \textbf{NCE}    & {\textbf{61.8}} & {\textbf{-32.9}} & 47.2 & {\textbf{89.7}} & {\textbf{100.0}} & 18.7 & 75.5 & {\textbf{98.7}} & 88.4 & 0.0 \\
\multicolumn{1}{l|}{} & \textbf{NCECE}  & 57.2 & -42.9 & 52.2 & 59.6 & 84.5 & 22.1 & 69.6 & 71.0 & 95.1 & 1.4 \\
\multicolumn{1}{l|}{} & \textbf{VR}     & 48.7 & -62.7 & {\textbf{52.5}} & 81.8 & 84.5 & 40.8 & 79.8 & 0.0 & {\textbf{97.2}} & 0.6 \\
\multicolumn{1}{l|}{\multirow{-8}{*}{\rotatebox{90}{\textbf{PointNet++}}}}
& \textbf{VRCE}   & 40.2 & -79.4 & 48.2 & 62.4 & 84.5 & 67.0 & 74.9 & 49.2 & 0.3 & 10.3 \\
\midrule
\multicolumn{1}{l|}{} & \textbf{CE}     & {\textbf{51.1}} & \textbf{-56.6} & 48.1 & 56.3 & {\textbf{95.8}} & 50.2 & 76.7 & 41.3 & 13.4 & 71.2 \\
\multicolumn{1}{l|}{} & \textbf{DR}     & 44.9 & -63.7 & 43.8 & 51.9 & 80.3 & 67.9 & 72.6 & 0.0 & 0.0 & {\textbf{100.0}} \\
\multicolumn{1}{l|}{} & \textbf{ARPL}   & 48.2 & -65.4 & {\textbf{56.5}} & 73.6 & 87.3 & 27.7 & {\textbf{73.5}} & 2.4 & 15.5 & 80.9 \\
\multicolumn{1}{l|}{} & \textbf{SupCon} & 38.7 & -78.0 & 45.7 & 40.0 & 69.0 & 65.7 & 72.0 & 47.1 & 18.5 & 0.1 \\
\cmidrule{2-12}
\multicolumn{1}{l|}{} & \textbf{NCE}    & 44.8 & -65.6 & 51.6 & 46.8 & 32.4 & {\textbf{75.7}} & 72.3 & {\textbf{93.7}} & 1.1 & 0.0 \\
\multicolumn{1}{l|}{} & \textbf{NCECE}  & 41.4 & -72.5 & 49.0 & 55.1 & 29.6 & 60.1 & 71.5 & 56.1 & 1.4 & 22.0 \\
\multicolumn{1}{l|}{} & \textbf{VR}     & 44.5 & -61.5 & 45.0 & 51.4 & 71.8 & 62.6 & 69.0 & 0.3 & {\textbf{99.4}} & 1.7 \\
\multicolumn{1}{l|}{\multirow{-8}{*}{\rotatebox{90}{\textbf{Point Transformer}}}}
& \textbf{VRCE}   & 45.2 & -68.8 & 53.0 & {\textbf{75.9}} & 73.2 & 12.2 & 73.8 & 49.2 & 29.5 & 10.3 \\
\bottomrule
\end{tabularx}}
\label{tab:argo2_ewc}
\end{table*}

\end{document}